%% file: egpaper.tex
\documentclass[10pt,twocolumn,letterpaper]{article}
\pdfoutput=1
\usepackage{wacv}
\usepackage{times}
\usepackage{epsfig}
\usepackage{graphicx}
\usepackage{amsmath}
\usepackage{amssymb}
\usepackage{booktabs}
\usepackage{mathtools}
\usepackage{microtype}
\usepackage[table,xcdraw]{xcolor}
\usepackage{caption}
\usepackage{placeins}
\usepackage{color, colortbl}
\usepackage{stfloats}
\usepackage{enumitem}
\usepackage{tabularx}
\usepackage{xstring}
\usepackage{multirow}
\usepackage{xspace}
\usepackage{url}
\usepackage{subcaption}
\usepackage[hang,flushmargin]{footmisc}
\usepackage{floatrow}
\usepackage{array,multicol,booktabs,tabularx}
\usepackage{multirow}
\usepackage[accsupp]{axessibility}
\usepackage{blindtext}

\usepackage{algorithm2e}
\RestyleAlgo{ruled}
\SetKwComment{Comment}{/* }{ */}

\DeclareMathOperator{\softmax}{softmax}
\DeclareMathOperator{\Seq}{Seq}
\DeclareMathOperator{\LSTM}{LSTM}
\DeclareMathOperator{\bsi}{beam-search-inference}

\def\wacvPaperID{755} 

\wacvalgorithmstrack   

\wacvfinalcopy 

\ifwacvfinal
\usepackage[breaklinks=true,bookmarks=false]{hyperref}
\else
\usepackage[pagebackref=true,breaklinks=true,colorlinks,bookmarks=false]{hyperref}
\fi

\begin{document}

\title{Seq-UPS: Sequential Uncertainty-aware Pseudo-label Selection for \\Semi-Supervised Text Recognition}

\author{Gaurav Patel \qquad Jan Allebach \qquad Qiang Qiu \\
School of Electrical and Computer Engineering, Purdue University, USA\\
{\tt\small \{gpatel10, allebach, qqiu\}@purdue.edu}}

\maketitle
\thispagestyle{empty}

\begin{abstract}
  This paper looks at semi-supervised learning (SSL) for image-based text recognition. One of the most popular SSL approaches is pseudo-labeling (PL). PL approaches assign labels to unlabeled data before re-training the model with a combination of labeled and pseudo-labeled data. However, PL methods are severely degraded by noise and are prone to over-fitting to noisy labels, due to the inclusion of erroneous high confidence pseudo-labels generated from poorly calibrated models, thus, rendering threshold-based selection ineffective. Moreover, the combinatorial complexity of the hypothesis space and the error accumulation due to multiple incorrect autoregressive steps posit pseudo-labeling challenging for sequence models. To this end, we propose a pseudo-label generation and an uncertainty-based data selection framework for semi-supervised text recognition. We first use Beam-Search inference to yield highly probable hypotheses to assign pseudo-labels to the unlabelled examples. Then we adopt an ensemble of models, sampled by applying dropout, to obtain a robust estimate of the uncertainty associated with the prediction, considering both the character-level and word-level predictive distribution to select good quality pseudo-labels.  Extensive experiments on several benchmark handwriting and scene-text datasets show that our method outperforms the baseline approaches and the previous state-of-the-art semi-supervised text-recognition methods.
\end{abstract}
\section{Introduction}

\label{sec:intro}
\begin{figure}[ht]
    \centering
    \includegraphics[width=0.5\linewidth]{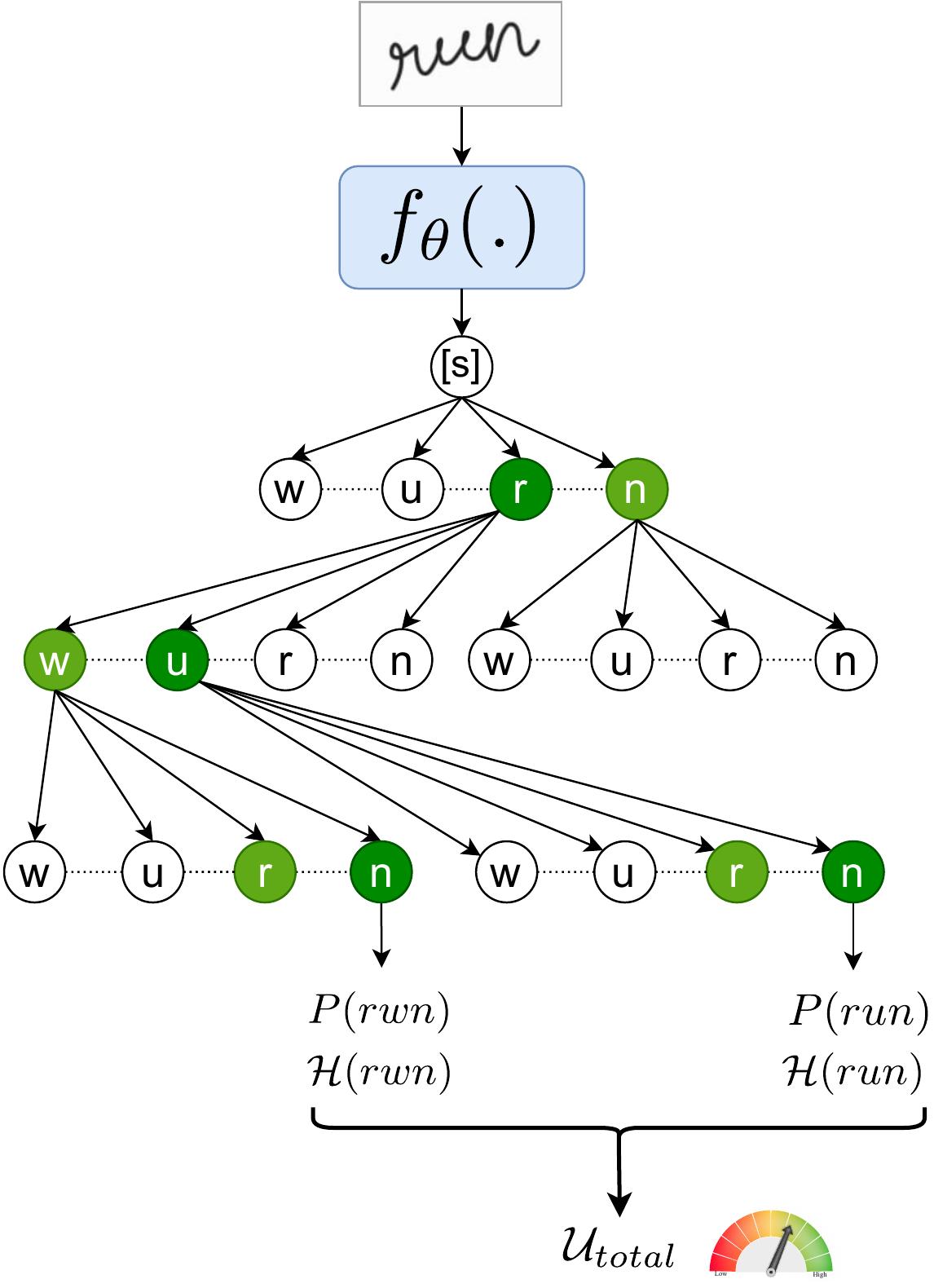}
    \caption{An overview of Beam-Search inference (beam width = 2) on recognizing an unlabeled text-image and populating the hypotheses set. The framework considers all the accumulated hypotheses to approximate the \emph{total uncertainty} ($\mathcal{U}_{total}$) by importance-sampling.}
    \label{fig:beam_search}
\end{figure}

Text recognition has garnered a great deal of attention in recent times \cite{strera}, owing primarily to its commercial applications. Since the introduction of deep learning, great strides \cite{deepSTR1,deepSTR5,deepSTR11,deepSTR8,deepSTR2,deepSTR9,deepSTR6,deepSTR7,deepSTR3,deepSTR10,deepSTR4} have been made in recognition accuracy on various publicly available benchmark datasets \cite{ICDAR15,ICDAR13,CVL,ICDAR03,IAM,IIIT,SVTP,CUTE,SVT}. These models, however, are heavily reliant on a large amount of labeled data with complete character-sequence as labels, which is laborious to obtain. Aside from fully-supervised text recognition, very few attempts have been made to utilize unlabelled data samples to improve the model's performance \cite{seqCLR,domain2,deepHTR4}.

Semi-supervised learning paradigms have been developed to address the preceding issues, and primarily pseudo-label-based semi-supervised learning (PL-SSL) methods have caught much attention. In a PL-SSL configuration, a smaller labeled set is used to train an initial seed model and then applied to a larger amount of unlabeled data to generate hypotheses. Furthermore, the unlabeled data along with its most reliable hypothesis as the label is combined with the training data for re-training, this methodology utilizes both the labeled and unlabelled data-points to re-train the complete model, allowing the entire network to exploit the latent knowledge from unlabelled data-points as well. However, on the other hand, PL-SSL is sensitive to the quality of the selected pseudo-labels and suffers due to the inclusion of erroneous highly confident pseudo-labels generated from poorly calibrated models, resulting in noisy training \cite{UPS}. Moreover, for image-based text recognition, that requires predictions of characters at each time-step for each input image, pseudo-labeling is much more challenging due to the combinatorial vastness of the hypotheses space and the fact that a single incorrect character prediction renders the entire predicted sequence false. Additionally, in the PL-SSL setup, handling erroneous predictions from the model trained with a relatively small amount of labeled data, and being able to exclude them in the beginning of the training cycle is highly essential. Therefore, correct hypotheses generation and selection are of massive importance in such a framework.

This work proposes a reliable hypotheses generation and selection method for PL-SSL for image-based text recognition. We suggest a way to estimate the uncertainty associated with the prediction of character sequences for an input image that gives a firm estimate of the reliability of the pseudo-label for an unlabelled data sample and then based on the estimated uncertainty, select the examples which have a low likelihood of an incorrectly assigned pseduo-label. Our methodology stems from the two primary observations that suggest (a) for pseudo-labeling based SSL schemes, choosing predictions with low uncertainty reduces the effect of poor calibration, thus improving generalization \cite{UPS} and (b) a high positive correlation exists between the predictive uncertainty and the token error rate, for a deep neural network- based language model, suggesting if the model produces a high uncertainty for an input image, its highly likely that the prediction used as the pseudo-label is incorrect \cite{u_wer}. 

Nevertheless, the majority of the unsupervised uncertainty estimation methods have concentrated on conventional unstructured prediction tasks, such as image classification and segmentation. Uncertainty estimation of an input image for text recognition, inherently a sequence prediction task, is highly non-trivial and poses various challenges \cite{uncertainty-estimate}: (a) Recognition models do not directly generate a distribution over an infinite set of variable-length sequences, and (b) an auto-regressive sequence prediction task, such as text recognition, does not have a fixed hypotheses set; therefore, debarring expectation computation on the same.

To circumvent these challenges we use Beam-Search inference (Figure \ref{fig:beam_search}) to yield high probability hypotheses for each unlabelled data-point on the seed model trained with the given labeled data. Moreover, the hypotheses obtained are used to approximate the predictive distribution and obtain expectations over the set. We term this process as \textit{deterministic inference}, which generates definite and a distinct hypotheses set for each image that aid in approximating the predictive distribution. Furthermore, to compute the uncertainty associated with an input, we take a \emph{Bayesian} approach to ensembles as it produces an elegant, probabilistic and interpretable uncertainty estimates \cite{uncertainty-estimate}.

We use \emph{Monte-Carlo-Dropout} (MC-Dropout) \cite{mcdropout}, which alleviates the need to train multiple models simultaneously and allows us to utilize Dropout to virtually generate multiple models (with different neurons dropped out of the original model) as Monte-Carlo samples and perform inferences on the sampled models, on each of the sequences in the hypotheses set by \textit{teacher-forcing} \cite{teacher-forcing}, terming it as \textit{stochastic inference}. Our motivation to utilize \textit{teacher-forcing} in the pseudo-labeling phase is  to enforce prediction consistency across all the sampled models in the ensemble such that we can estimate the predictive distribution of every hypothesis obtained with \textit{deterministic-inference}. Finally, the predictive posterior for each hypothesis is obtained by taking the expectation over all the sampled models. Furthermore, the obtained predictive posterior is used to compute an information-theoretic estimate of the uncertainty, which estimates the \emph{total uncertainty} \cite{MCdropout_robust}, considering both the character-level and word-level predictive posteriors and serves as a robust selection criterion for the pseudo-labels. Figure \ref{fig:beam_search} shows an intuitive idea behind Beam-Search inference to generate multiple hypotheses for a normalized uncertainty estimate. Finally, we test our method on several handwriting and scene-text datasets, comparing its performance to state-of-the-art text-recognition methods in semi-supervised setting. Moreover, we demonstrate the robustness of our uncertainty estimation using \emph{prediction rejection curves} \cite{rejection_curve_2,rejection_curve_1} based on the  Word Error Rate (WER).

In sum the key-points are: (a) We propose an uncertainty-aware pseudo-label-based semi-supervised learning framework, that utilizes Beam-Search inference for pseudo-label assignment, and a character and sequence aware uncertainty estimate for sample selection. (b) We utilize \textit{teacher-forcing} \cite{teacher-forcing}, primarily employed to train sequence-models, in the pseudo-labeling phase to enforce prediction consistency across all the sampled models in the ensemble, to estimate the predictive distribution.(c) Finally, the methods are evaluated on several challenging handwriting and scene-text datasets in the SSL setting.
    

\section{Related Work}
\noindent\textbf{Text Recognition: }Attention-based sequential decoders \cite{attention} have become the cutting-edge framework for text recognition in scene-text \cite{deepSTR2,deepSTR3,deepSTR12,deepSTR13} and handwriting \cite{deepHTR1,deepHTR2,deepHTR3,deepHTR4}. Besides, various incremental propositions \cite{deepSTR5,deepSTR15,deepSTR14} have been made, such as introducing or improving the rectification module \cite{deepSTR9,deepSTR7,deepSTR12,deepSTR13}, designing a multi-directional convolutional feature extractor \cite{deepSTR16}, enhancing the attention mechanism \cite{deepSTR17,deepSTR18}, and stacking multiple Bi-LSTM layers for better context modeling \cite{deepSTR17,deepSTR6,deepSTR7}. With all the diversity present in text recognition models, the seminal work by Baek \textit{et al.} \cite{deeptext} provides a unifying framework for text recognition that furnishes a modular viewpoint for existing methods by suggesting that the recognition system be divided into four distinct stages of operations, namely: (a) spatial-transformation (Trans.), (b) feature extraction (Feat.), (c) sequence modeling (Seq.), and (d) prediction (Pred.). Furthermore, the framework provides not only the existing methods but also their possible variants \cite{deeptext}, and demonstrates that most of the state-of-the-art \cite{deepSTR11,deepSTR8,deepSTR9,deepSTR6,deepSTR7,deepSTR10} methods fall under this framework.\\
\textbf{Semi-Supervised Learning: }When labels are scarce or expensive to obtain, semi-supervised learning provides a powerful framework to exploit unlabeled data. SSL algorithms based on deep learning have demonstrated to be effective on standard benchmark tasks. Numerous semi-supervised learning works have been published in recent years, including consistency regularization-based methods \cite{consistency1,consistency4,consistency2,consistency3}, label propagation \cite{label_prop_2,label_prop_1}, self-training \cite{self_training_2,SL-SSL,UPS}, data augmentation \cite{dataaug2,dataaug1} and entropy regularization \cite{entropy_reg}. Even though semi-supervised learning is rapidly evolving, it is commonly employed and explored for unstructured predictions tasks \textit{e.g.} classification or semantic segmentation. However, text recognition is an structured prediction task and therefore, off-the-shelf semi-supervised methods are unlikely to be directly applicable for the current use case.\\
\textbf{Semi-Supervised Text-Recognition: }Notwithstanding the obvious benefits, most text recognition systems do not currently use unlabeled text images. Handwriting recognition in particular, is usually based on fully-supervised training \cite{deepHTR6,deepHTR5}, whereas scene-text models are primarily trained on synthetic data \cite{ST,MJ}. However, Kang \textit{et al.} \cite{domain2} and Zhang \textit{et al.} \cite{deepHTR4}  recently proposed domain adaptation techniques for using an unlabeled dataset alongside labeled data. Nevertheless, they introduce specific modules to impose domain alignment. Fogel \textit{et al.} \cite{unsup2} proposed a completely unsupervised scheme for handwritten text images, in which a discriminator forces predictions to align with a distribution of a given text corpus. Nonetheless, these methods necessitate restricting the recognizer to use only local predictions. Moreover, Gao \textit{et al.} \cite{SSL-STR} concoct a reward-based methodology in a reinforcement-learning framework to perform SSL for scene-text recognition, however, they introduce an additional embedding network to compute the embedding distance to formulate the reward function, thus introducing a computational overhead at the time of training. Additionally, Aberdam \textit{et al.} \cite{seqCLR} propose a \textit{sequence-to-sequence} contrastive self-supervised learning framework for text recognition, which implicitly shows SSL capabilities due to the generalization power of self-supervised learners \cite{SL-SSL}, the method learns generalized image features with self-supervised pre-training, which is leveraged to initialize the weights at the fine-tuning phase. Moreover, we hypothesize that the model's performance can further be improved at the fine-tuning phase by utilizing the unlabeled data.
\label{sec:related}


\section{Methodology}
\label{sec:method}

\begin{figure}
\centering
\includegraphics[width=0.98\linewidth]{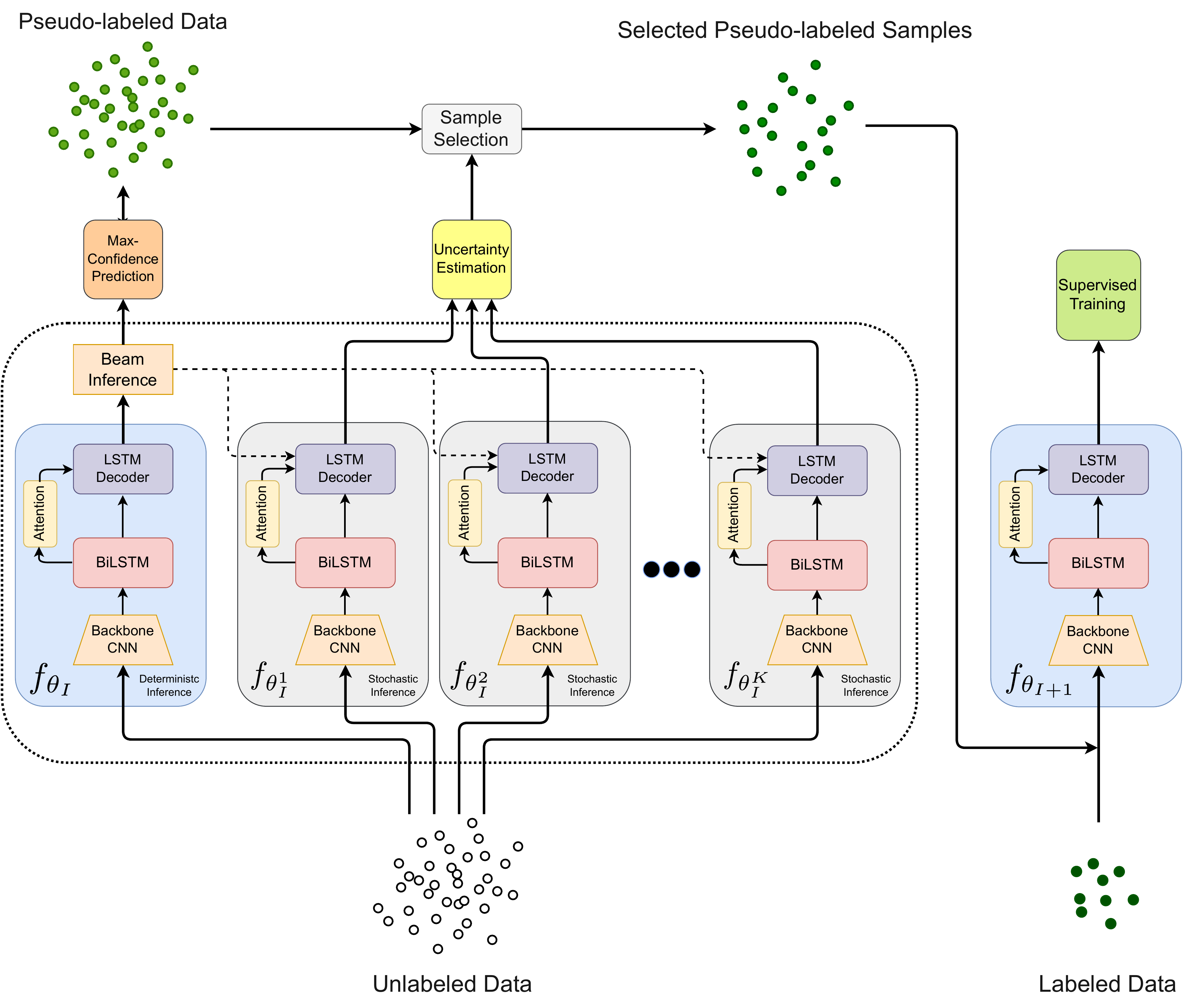}
\caption{Schematic diagram representing pseudo label generation and uncertainty-based label selection strategy. The figure depicts a snap-shot of the pseudo-labeling process at the end of the $I$-th training iteration. The unlabeled data points in $\mathcal{D}_{u}$ are assigned pseudo-labels by the \emph{deterministic inference}. Furthermore, uncertainty associated with the each unlabelled sample is computed by \emph{stochastic-inference} with $K$ ensembles; and then samples with low uncertainty values, based on a threshold parameter, are selected to be included in the training set $\mathcal{D}_{train}$ along with $\mathcal{D}_{l}$ for the $(I+1)$-th training iteration. The dotted arrows represent the \emph{teach-forcing} \cite{teacher-forcing} applied at the time of uncertainty estimation. (The spatial-transformation module is omitted for brevity.) }\label{fig:main}
\end{figure}

We start with a preliminary introduction describing the formulation of a generic text recognition model in \ref{preliminary}. Then, we describe the semi-supervised framework as shown in Figure \ref{fig:main} in subsection \ref{ssl}  and describe the proposed sequential uncertainty estimation methodology in \ref{ue}. Lastly in subsection \ref{rue} we verify the robustness of the introduced uncertainty estimation and show the estimated uncertainty's correlation with word error rate (WER).
\subsection{Preliminary}
\label{preliminary}
The text recognition model $f_{\boldsymbol{\theta}}(.)$ attempts to predict the machine readable character sequence $Y$ given an image $X$. Among the numerous text-recognition models with different backbones and sequential decoding schemes, we adopt a generic recognition model with an attention based sequential decoding scheme \cite{deeptext}. Most state-of-the-art text-recognition choices dealing with irregular or handwritten texts are derivatives of our chosen model. Our text recognition model consists of primarily four components: (a) a Spatial Transformer Network \cite{STN} to ease downstream stages by image normalization, (b) a backbone convolutional feature extractor, (c) a contextual sequence modelling stage, and (d) a sequence model that autoregressively predicts the characters at each time step. Let the extracted convolutional feature map be $V = \{v_{i}|v_{i} \in \mathbb{R}^{C}\}_{i=1}^{W}$, where $W$ is the number of columns in the feature map and $C$ be the number of channels. The feature map $V$ passes through a sequence model, denoted as $\Seq(.)$  to generate a contextual sequence $H$, i.e., $H = \Seq(V) = \{ h_{i}|h_{i} \in \mathbb{R}^{D} \}_{i=1}^{W}$, where $D$ is the hidden-state dimension. Furthermore, at the $t$-th step, the decoder predicts a probability mass across the character space, $p(y_{t}) = \softmax(W_{0}s_{t} + b_{0})$, where $W_{0}$ and $b_{0}$ are trainable parameters, and $s_{t}$ is the decoder LSTM hidden state at step $t$ defined as $s_{t}=\LSTM(y_{t-1}, c_{t}, s_{t-1})$. Here, $c_{t}$ is a context vector computed as a weighted sum of the elements in $H$ from the previous step, defined as $c_{t} = \sum_{i=1}^{I}{\alpha_{t,i}h_{i}}$, such that $\alpha_{t,i} = \frac{\exp{e_{t,i}}}{\sum_{k}\exp{e_{t,k}}}$, with $e_{t,i} = w_{a}^{\top}\tanh(W_{b}s_{t-1} + W_{c}h_{i} + b)$, where $w_{a}$, $W_{b}$, $W_{c}$, and $b$ are trainable parameters. The complete recognition model is trained end-to-end using cross-entropy loss $\mathcal{CE}(·, ·)$ summed over the ground-truth sequence $Y = \{y_{t} \}_{t=1}^{S}$, where $y_{i}$ is the one-hot encoded vector in $\mathbb{R}^{E}$, where $E$ is the character vocabulary size and $S$ denotes the sequence length.
\begin{equation}
    \mathcal{L}_{CE} = \sum_{t=1}^{S}{\mathcal{CE}(y_t, p(y_t))} = \sum_{t=1}^{S}\sum_{i=1}^{E}y_{t,i}\log(p(y_{t,i})).
\end{equation}
Moreover, due to the autoregressive nature of the LSTM decoder, we can factorize the distribution over $Y$ as product of conditional distributions, conditioned on previously predicted outputs $Y_{<t} = \{y_{i}\}_{i=t-1}^{1}$, the input $X$ and all the trainable parameters denoted with $\boldsymbol{\theta}$ as,

\begin{equation}
P(Y|X, \boldsymbol{\theta}) = \prod_{t=1}^{S}P({y}_{t}|Y_{<t}, X,\boldsymbol{\theta}).
\end{equation}

\subsection{Pseudo-label Assignment and Selection for Semi-Supervised Text Recognition}
\label{ssl}
Here we present an overview of our uncertainty-based pseudo-label semi-supervised framework, as shown in Figure \ref{fig:main}. Let us denote the labeled dataset containing $N_{l}$ samples as $\mathcal{D}_{l} = \{X^{l}_{i},Y^{l}_{i}\}_{i=1}^{N_{l}}$, where $X_{i}^{l}$ and $Y_{i}^{l}$ denote the $i$-th training image and its corresponding character-sequence label, and let $N_{u}$ denote the number of images that are presented in the unlabelled set denoted as $\mathcal{D}_{u}=\{X_{i}^{u}\}_{i=1}^{N_{u}}$, not associated with any ground-truth. In the proposed pseudo-labeling method, the recognition network undergoes multiple rounds of training. In the initial round, the model is first trained in a supervised manner with the labeled dataset, \textit{i.e.} $\mathcal{D}_{train} = \mathcal{D}_{l}$. Furthermore, the trained model is used to estimate labels for unlabeled data as $\tilde{\mathcal{D}}_{u}=\{X_{i}^{u}, \tilde{Y}_{i}^{u}\}_{i=1}^{N_{u}}$, where $\tilde{Y}_{i}^{u}$ denotes the pseudo-label for the $i$-th sample from the unlabeled pool. Concretely, given an unlabeled image $X_{i}^{u}$, Beam-Search inference is performed on the trained recognition network parameterized by $\boldsymbol{\theta}$ to obtain a set of top-$B$ high confidence predictions denoted by $\mathcal{B}_{i} = \{Y^{(b)}_{i}\}_{b=1}^{B}$ and is assigned the pseudo-label $\tilde{Y}_{i}^{u} = \operatorname*{arg\,max}_{Y^{(b)}_{i}}\{P(Y^{(b)}_{i}|X_{i}^{u},\boldsymbol{\theta})\}_{b=1}^{B}$. Then we compute the uncertainty, which in its functional form can be denoted as $\mathcal{U}(X_{i}^{u}, \mathcal{B}_{i})$, associated with the predictions from the input $X_{i}^{u}$. After this, the pseudo-label selection process is performed. Let $\boldsymbol{q} = \{q_{i}\}_{i=1}^{N_{u}}, q_{i} \in \{0,1\}$, denote a binary vector, where a value of $1$ at the $i$-th location depicts the selection of that particular unlabeled sample along with its pseudo label for subsequent training.

\begin{equation}
   q_{i}=\begin{cases}
1, & \text{if} \ \mathcal{U}(X_{i}^{u}, \mathcal{B}_{i}) \leq \tau,\\
0, & \text{otherwise},
\end{cases}
\end{equation} where $\tau$ is the selection threshold hyperparameter. Unlabeled samples with low uncertainty are selected and included in the training dataset $\mathcal{D}_{train} = \mathcal{D}_{l} \cup \{ \{X_{i}^{u}, \tilde{Y}_{i}^{u}\}_{i=1}^{N_{u}}| \text{if} \ q_{i}=1 \}$ for re-training the recognition model.

\subsection{Uncertainty Estimation for Sequential Prediction}
\label{ue}
Now we elaborate on the adopted uncertainty estimation methodology for pseudo-label selection. We take an ensemble-based approach for uncertainty estimation, also referred to as MC-Dropout \cite{mcdropout}. Consider an ensemble of models with the predicted probability distribution $\{P(Y|X,\boldsymbol{\theta}^{k})\}_{k=1}^{K}$, with each ensemble model parameter denoted by $\boldsymbol{{\theta}}^k$, where $K$ denotes the number of ensembles, sampled from a posterior $\pi_{p}(\boldsymbol{\theta})$. Here, $\pi_{p}(\boldsymbol{\theta})$ is the set of all possible parameters with dropout applied to it with a given probability $p$. Moreover, the predictive posterior is defined as the expectation over the ensemble. In our case, to approximate the predictive posterior, we generate an ensemble of models by applying dropout, and virtually generate $K$ versions from the same model, and define the predictive distribution as,
\begin{equation}
\begin{aligned}
P(Y|X) \coloneqq \int{P(Y|X, \boldsymbol{\theta})\pi(\boldsymbol{\theta})d\boldsymbol{\theta}} \simeq \frac{1}{K} \sum_{k=1}^{K}{P(Y|X, \boldsymbol{\theta}^{k})},\\ \boldsymbol{\theta}^{k} \sim \pi_{p}(\boldsymbol{\theta}).
\label{eq:1}
\end{aligned}
\end{equation}
Furthermore, the \textit{total uncertainty} \cite{MCdropout_robust} associated with the prediction of $Y$ given an input $X$, is given by the entropy of the predictive distribution,
\begin{equation}
\begin{aligned}
\mathcal{H}[P(Y|X)] & \coloneqq \mathbb{E}_{P(Y|X)}[-\ln{P(Y|X)}]\\ 
& = - \sum_{Y\in\mathcal{Y}}{P(Y|X)}\ln{P(Y|X)},
\label{eq:2}
\end{aligned}
\end{equation} where $\mathcal{Y}$ denotes set of all possible hypotheses.
However, the expression in (\ref{eq:2}) is not tractable for sequence models due to the ungovernable combinatorial vastness of the hypothesis space. Therefore, to devise a robust estimate of uncertainty for text recognition, we define a character and sequence aware uncertainty estimate that is tractable. For each unlabeled example, we use the hypotheses present in the set $\mathcal{B}_{i}$ which contains the  top-$B$ high-probability predicted sequences obtained with Beam-Search inference at the time of decoding. Note that we had already obtained high-confidence hypotheses using \emph{deterministic inference} at the time of pseudo label assignment, as described previously in \ref{ssl}. Furthermore, we perform inference on the different instances of the same model obtained with dropout by \emph{teacher-forcing} to calculate the uncertainty measure as described below,
\begin{align}
    \nonumber
    \mathcal{U}(X_{i}, \mathcal{B}_{i}) & = -\sum_{b=1}^{B} \sum_{t=1}^{L_{b}} \frac{\omega_{b}}{L_{b}}\mathcal{H}[P(y_{t}|Y^{(b)}_{<t}, X_{i})],\\ 
    \omega_{b} & = \frac{\exp({\frac{1}{T}\ln{P(Y^{(b)}|X_{i}}))}}{\sum_{b=1}^{B}\exp({\frac{1}{T}\ln{P(Y^{(b)}|X_{i}}))}},
    \label{eq:6}
\end{align}
where $L_{b}$ denotes the length of the hypothesis and $T$ denotes the temperature hyperparameter. Additionally, for the sequence-level measures of uncertainty, they need to be adjusted such that uncertainty associated with each hypothesis $Y^{(b)}$ within the hypotheses set $\mathcal{B}_{i}$ is weighted in proportion to its probability (confidence) \cite{active_handwriting}, denoted by $\omega_{b}$. 
The defined uncertainty estimate takes into consideration the uncertainty at every character prediction and all the word sequences present in the hypotheses set $\mathcal{B}_{i}$. Thus, we approximate the entire hypotheses space with a few highly probable predictions. Moreover, our motivation to use \textit{teacher-forcing} is to impose prediction consistency across all the models in the ensemble. Given an input image to the ensemble, the hypotheses set generated by each model is not guaranteed to be the same, therefore, we \textit{teacher-force} the hypotheses in $\mathcal{B}_{i}$ to infer from the models sampled for the ensemble.

\subsection{Robust Uncertainty Estimation?}
\label{rue}
\begin{figure}[ht]
    \centering
\includegraphics[width=0.99\linewidth]{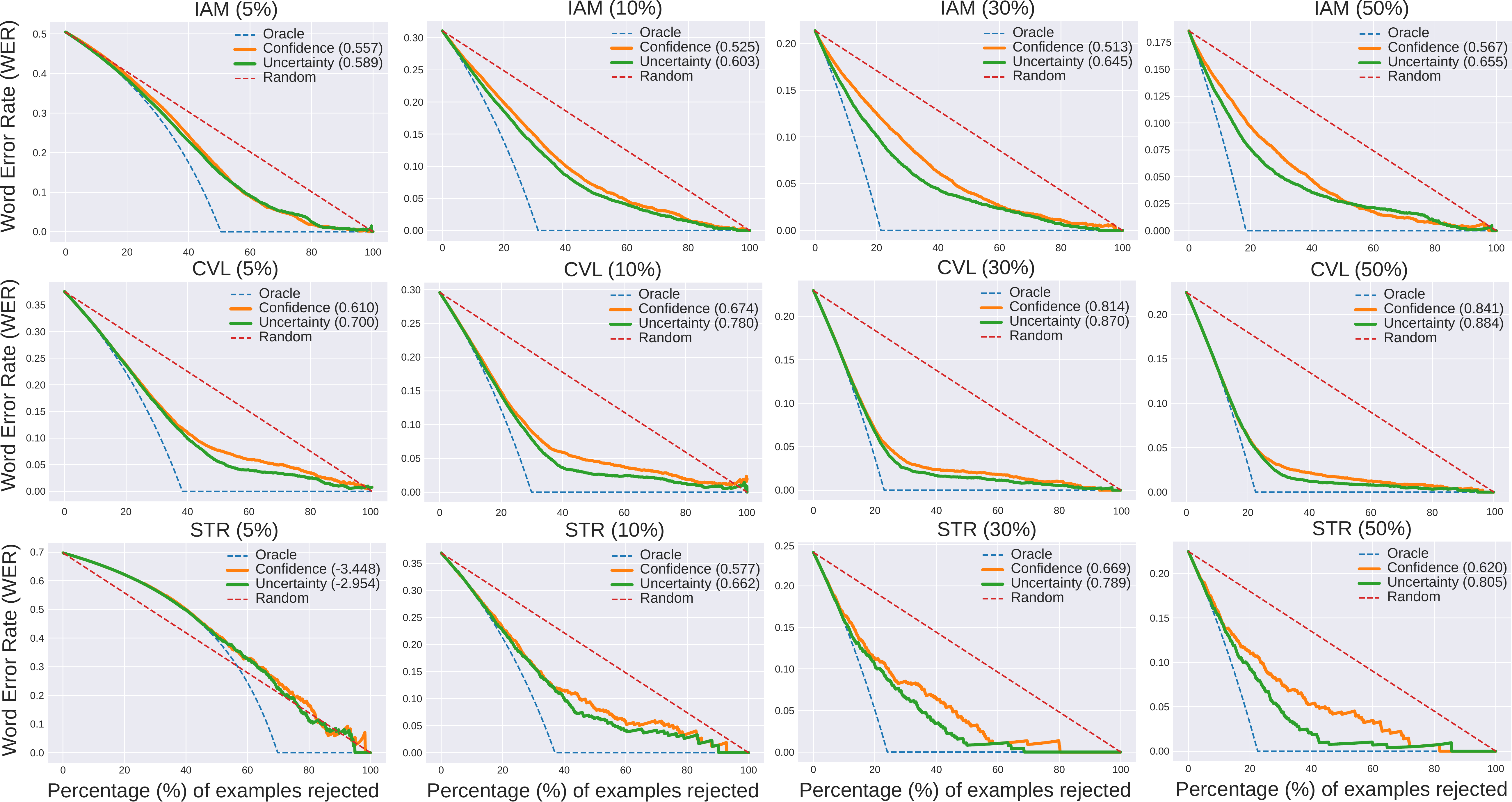}
     \caption{Prediction Rejection Curves. Values in parenthesis in the legend field represent the \textit{PRR} of the corresponding curve.}
     \label{fig:rej_curves}
\end{figure}

Ideally, we would like to identify all of the inputs that the model mis-predicts based on its uncertainty measure, and then we may decide to either include or choose not to use the predictions as pseudo-labels. Malinin \textit{et al.} \cite{rejection_curve_2,uncertainty-estimate,rejection_curve_1} proposed using the \emph{prediction rejection curves} and the \emph{prediction rejection area ratios} to qualitatively and quantitatively assess an uncertainty measure's correlation to misclassifications, respectively. Instead of considering misclassifications, we examine our uncertainty measure's correlation to Word Error Rate (WER) = $1 - \frac{word\ prediction\ accuracy}{100}$ (Figure \ref{fig:rej_curves}). Given that, if the error metric is a linear function of individual errors, the expected \emph{random} rejection curve should be a straight line from the base error rate on the left to the lower right corner if the uncertainty estimates are uninformative. However, if the estimates of uncertainty are \emph{perfect} and always larger for an incorrect prediction than for a correct prediction, then the \emph{oracle} rejection curve would be produced. When at a percentage of rejected examples equal to the number of false predictions, the \emph{oracle} curve will fall linearly to 0 error.
A rejection curve produced by imperfect but informative uncertainty estimates will rest between the \emph{random} and \emph{oracle} curves.
The area between the selected measure's rejection curves and \emph{random} curve $AR_{r}$ and the area between the \emph{oracle} and \emph{random} curves $AR_{orc}$ can be used to assess the quality of the measure's rejection curve. \emph{prediction rejection area ratio} (\textit{PRR}) is defined as $PRR = \frac{AR_{r}}{AR_{orc}}.$ 

Figure \ref{fig:rej_curves}  depicts the rejection curves of the baseline text recognition model trained on different portions of labeled data on the handwriting and the scene-text datasets. We show the rejection curve comparison on two measures: (a) Confidence of the most probable hypothesis (in Blue) and (b) Our proposed sequence uncertainty measure (in Orange). For the confidence curves, the examples are rejected based on their confidence values, low-to-high, and for the uncertainty curves based on their uncertainty values, high-to-low. The rejection curves substantiate the superiority of our uncertainty-based pseudo-label selection compared to confidence-based selection. A rejection area ratio of $1$ represents optimal rejection, while $0$ represents \emph{random} rejection. In addition, the metric is independent of the model's performance, allowing it to be used to compare models with different base error rates. The higher \textit{PRR} values for uncertainty-based rejection suggest a stronger correlation to WER. Primarily, in Figure \ref{fig:rej_curves}, we observe that Uncertainty-\emph{PRR} in all the cases is greater than Confidence-\emph{PRR}, validating the uncertainty-measure's stronger correlation, conveying that the estimated uncertainty is a better estimation of the WER compared to confidence of the prediction. Especially, the gap in the \textit{PRR} values is more significant for models trained with a lower fraction (5\% and 10\%) of labeled data. Additionally in the supplemental material we provide the prediction rejection curves w.r.t the \textit{character error rate}.


\section{Experiments}
\label{sec:experiments}


\subsection{Experimental Settings}
\paragraph{Datasets:}
\begin{figure}[ht]
    \centering
    \includegraphics[width=0.94\linewidth]{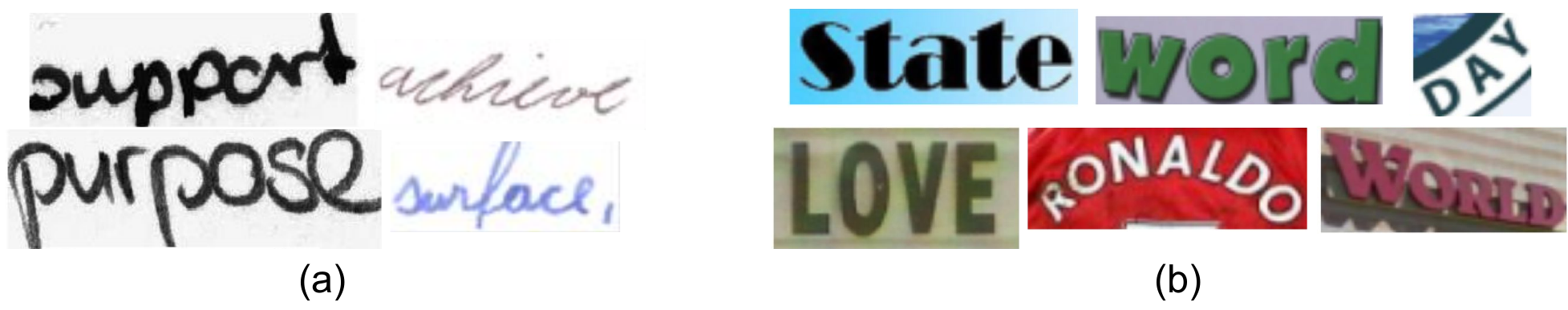}
    \caption{Examples of (a) handwritten-text images (IAM and CVL) and (b) scene-text images (IIIT5K, SVT, IC13, IC15, SVTP and CUTE)}
    \label{fig:text_examples}
\end{figure}
We compare the performances of the methods on a variety of publicly available datasets for handwriting and scene-text recognition. We consider the IAM \cite{IAM} and CVL \cite{CVL} word-level datasets for handwriting, and the datasets: IIIT5k \cite{IIIT}, SVT \cite{SVT}, IC13 \cite{ICDAR13}, IC15 \cite{ICDAR15}, SVTP \cite{SVTP}, and CUTE \cite{CUTE} for scene-texts as shown in Figure \ref{fig:text_examples}. To train the handwriting recognition models, we use the training data split of the IAM dataset and the CVL dataset. For scene-text recognition, unlike the previous works \cite{seqCLR,deeptext} that use a large amount of synthetic training data \cite{ST,MJ} to train scene-text models, we use a combination of multiple real scene-text recognition datasets \cite{RealSTR_data}. More details can be found in the supplemental material.

\paragraph{Training Details:}
We use the best performing model configuration introduced by Baek \textit{et al.} \cite{deeptext} commonly dubbed as TRBA (\textbf{T}PS-\textbf{R}esidual-\textbf{B}iLSTM-\textbf{A}ttention) with dropout layers between the convolution layers. We pre-resize all images to 32$\times$100 for training and testing and use 95 symbol classes: 52 case-sensitive letters, 10 digits, and 33 special characters. For the MC-Dropout \cite{mcdropout} ensembling we use a dropout probability of $0.1$ and $K=5$ ensembles. Furthermore, we use a beam-width of $5$ with the temperature value $T$ set to $0.01$ to weigh the hypotheses according to their confidence. Moreover, the sample selection threshold $\tau$ for uncertainty is set to a low value of $0.01$ for all the experiments in handwriting and scene-text recognition. For all the experiments we perform 5 iterations of pseudo-label based self-training and report the results of the best performing iteration after the first pseudo-labeling step. Furthermore we would like to emphasize that all of the pseudo-labeling and uncertainty estimation is only carried out after the supervised model training, and therefore it does not introduce any computational overhead at the time of training, contrary to the work of Gao \textit{et al.} \cite{SSL-STR}, which simultaneously learn an embedding network with a similar memory footprint as the recognition network, for reward-based gradient estimation. More about the training details can be found in the supplemental material.

\paragraph{Evaluation Metrics:}
For scene-text datasets we report the \emph{Word Prediction Accuracy} following Baek \textit{et al.} \cite{deeptext} and for handwriting datasets, additionaly, we report \emph{Character Error Rate (CER)} which is based on Levenstein distance \cite{cer,deepHTR6}.

\subsection{Baseline and State-of-the-art Comparisons}
\setlength{\tabcolsep}{4pt}
\begin{table*}[t]
\begin{center}
\caption{Semi-Supervised Handwriting Recogniton Results: Word level accuracy (Acc) (\%) and  Character Error Rate (CER) comparisons of models trained with 5\% and 10\% of the total available labeled data. Best results highlighted in bold.}\label{table:handwriting}
\resizebox{0.66\textwidth}{!}{
\begin{tabular}{l|cccccccc}
\hline \noalign{\smallskip}
\multirow{4}{*}{Method}             & \multicolumn{4}{c}{IAM \cite{IAM}}                                                                                  & \multicolumn{4}{c}{CVL \cite{CVL}}                                                                               \\ \cline{2-9} \noalign{\smallskip}
                                    & \multicolumn{8}{c}{Fraction of labeled data}                                                                                                                                                                     \\
                                    & \multicolumn{2}{c}{5\%}                           & \multicolumn{2}{c|}{10\%}                            & \multicolumn{2}{c}{5\%}                           & \multicolumn{2}{c}{10\%}                          \\ \cline{2-9} \noalign{\smallskip}
                                    & \multicolumn{1}{c}{Acc $\uparrow$} & \multicolumn{1}{c}{CER $\downarrow$} & \multicolumn{1}{c}{Acc $\uparrow$} & \multicolumn{1}{c|}{CER $\downarrow$}   & \multicolumn{1}{c}{Acc $\uparrow$} & \multicolumn{1}{c}{CER $\downarrow$} & \multicolumn{1}{c}{Acc $\uparrow$} & \multicolumn{1}{c}{CER $\downarrow$} \\ \noalign{\smallskip} \hline \noalign{\smallskip}
Supervised Baseline                 & 49.63                   & 28.47                   & 68.00                   & \multicolumn{1}{c|}{11.63} & 62.28         & 25.20                  & 70.31                   & 23.62                   \\
Pseudo-Labeling Baseline                & 54.57                  & 24.64                   & 71.06                   & \multicolumn{1}{c|}{13.23} & 64.38         & 29.74                  & 73.18                   & 23.42                   \\\noalign{\smallskip} \hline \noalign{\smallskip}
SeqCLR (All-to-instance) \cite{seqCLR}            & 64.58                   & 11.63                  & 70.54                   & \multicolumn{1}{c|}{9.72}   & 63.10         & 29.40              & 70.51                   & 24.15                         \\
SeqCLR (Frame-to-instance) \cite{seqCLR}          & 64.81                   & 14.41                  & 71.64                   & \multicolumn{1}{c|}{9.55}   & 63.08         & 27.30                   & 70.63                   & 23.89                    \\
SeqCLR (Window-to-instance) \cite{seqCLR}         & 64.16                   & 14.06                  & 71.43                   & \multicolumn{1}{c|}{8.68}   & 63.14          & 27.56                     & 70.58                   & 22.84                       \\ \noalign{\smallskip} \hline \noalign{\smallskip}
\textbf{Ours}                                & 69.37                   & 11.98                  & 74.53                   & \multicolumn{1}{c|}{8.85}  & \textbf{74.90}           & 23.10                   & 75.37                   & 21.79                   \\
\textbf{Ours w/ SeqCLR (All-to-instance)}    & 73.60                   & 9.03                   & \textbf{77.11}                   & \multicolumn{1}{c|}{7.47}  & 72.41           & \textbf{19.95}                    & 75.67                   & 21.26                         \\
\textbf{Ours w/ SeqCLR (Frame-to-instance)}  & \textbf{73.87}                   & \textbf{8.33}                        & 76.95              & \multicolumn{1}{c|}{7.47}  & 72.36            & 22.84              & 75.75                   & 22.05                   \\
\textbf{Ours w/ SeqCLR (Window-to-instance)} & 73.37                   & 8.85                    & 76.90                   & \multicolumn{1}{c|}{\textbf{7.29}} & 72.23            & 22.31                    & \textbf{75.76}                   & \textbf{20.21}                        \\ \noalign{\smallskip} \hline \noalign{\smallskip}

\end{tabular}}
\end{center}
\end{table*}
\setlength{\tabcolsep}{1.4pt}

In Table \ref{table:handwriting} we compare the performance of the methods on handwriting recognition task in the semi-supervised setting, \textit{i.e.}, being trained only with a portion of the labeled training dataset (5\% and 10\%) on Word Prediction Accuracy (Acc) and the Character Error Rate (CER) on IAM \cite{IAM} and CVL \cite{CVL} word-level datasets. In addition to the supervised baseline we include the pseudo-labeling baseline, in which, we don not employ the threshold based unlabeled sample selection but select all the unlabeled samples along with their pseudo label for model retraining. We make comparisons with the 3 state-of-the-art variants of SeqCLR \cite{seqCLR} with the BiLSTM projection head, which primarily differs in the instance mapping strategy used at the time of pre-training. All SeqCLR variants are first pre-trained with all the training examples, \textit{i.e.}, $\mathcal{D}_{train} = \mathcal{D}_{l}\cup\mathcal{D}_{u}$, and then the pre-trained model is used as an initialization, and subsequently the entire model is fine-tuned with only a portion of labeled training examples in $\mathcal{D}_{train}$ \textit{i.e.} $\mathcal{D}_{l}$, in a fully-supervised manner. For our method, we use the same portions of the labeled split as used in the baseline and the prior arts and train from scratch starting with $\mathcal{D}_{train} = \mathcal{D}_{l}$. Post the supervised training with $\mathcal{D}_{train}$, pseudo-labeling is performed to update $\mathcal{D}_{train}$ and the model is re-trained from scratch with the updated $\mathcal{D}_{train}$. We observe that our method consistently outperforms the baseline and all the variants of SeqCLR in the semi-supervised setting, especially when using 5\% of the labeled data, with an average gain of $\approx 4.5\%$ and $\approx 2.9\%$ in accuracy, and around $\approx 2.5\%$ and $\approx 4.5\%$ reduction in CER for the IAM and CVL datasets, respectively. Note that even our vanilla PL-SSL method (without the self-supervised pre-training) performs significantly better than the prior arts. 


\setlength{\tabcolsep}{4pt}
\begin{table*}[t]
\begin{center}
\caption{Semi-Supervised Scene-Text Recognition Results: Word level accuracy (\%) comparisons of models trained with 5\% and 10\% of the total available labeled data. Best results highlighted in bold.}
\label{table:STR}
\resizebox{0.9\textwidth}{!}{
\begin{tabular}{l|clclclclclclclclclclclclclcl}
\hline \noalign{\smallskip}
\multirow{3}{*}{Method}             & \multicolumn{4}{c}{IIIT5k \cite{IIIT}}                               & \multicolumn{4}{c}{SVT \cite{SVT}}                                & \multicolumn{4}{c}{IC13 \cite{ICDAR13}}                               & \multicolumn{4}{c}{IC15 \cite{ICDAR15}}                               & \multicolumn{4}{c}{SVTP \cite{SVTP}}                               & \multicolumn{4}{c}{CUTE \cite{CUTE}}                             & \multicolumn{4}{c}{Total}                             \\ \cline{2-29} \noalign{\smallskip}
                                    & \multicolumn{28}{c}{Fraction of labeled data}                                                                                                                                                                                                                                                                                                                                                               \\
                                    & \multicolumn{2}{c}{5\%}   & \multicolumn{2}{c|}{10\%}  & \multicolumn{2}{c}{5\%}   & \multicolumn{2}{c|}{10\%}  & \multicolumn{2}{c}{5\%}   & \multicolumn{2}{c|}{10\%}  & \multicolumn{2}{c}{5\%}   & \multicolumn{2}{c|}{10\%}  & \multicolumn{2}{c}{5\%}   & \multicolumn{2}{c|}{10\%}  & \multicolumn{2}{c}{5\%}   & \multicolumn{2}{c|}{10\%}  & \multicolumn{2}{c}{5\%}   & \multicolumn{2}{c}{10\%}  \\
                                    \noalign{\smallskip} \hline \noalign{\smallskip}
Supervised Baseline                 
& \multicolumn{2}{c}{41.63} & \multicolumn{2}{c|}{75.60} & \multicolumn{2}{c}{33.39} & \multicolumn{2}{c|}{69.55} & \multicolumn{2}{c}{46.11} & \multicolumn{2}{c|}{78.52} & \multicolumn{2}{c}{26.43} & \multicolumn{2}{c|}{55.51} & \multicolumn{2}{c}{24.96} & \multicolumn{2}{c|}{54.88} & \multicolumn{2}{c}{23.26} & \multicolumn{2}{c|}{48.96} & \multicolumn{2}{c}{35.32} & \multicolumn{2}{c}{67.30} \\
Pseudo-Labeling Baseline                 
& \multicolumn{2}{c}{51.30} & \multicolumn{2}{c|}{76.57} & \multicolumn{2}{c}{42.50} & \multicolumn{2}{c|}{72.18} & \multicolumn{2}{c}{54.88} & \multicolumn{2}{c|}{80.89} & \multicolumn{2}{c}{35.35} & \multicolumn{2}{c|}{57.40} & \multicolumn{2}{c}{31.16} & \multicolumn{2}{c|}{57.52} & \multicolumn{2}{c}{30.21} & \multicolumn{2}{c|}{50.00} & \multicolumn{2}{c}{44.25} & \multicolumn{2}{c}{69.02} \\\noalign{\smallskip} \hline \noalign{\smallskip}
Gao \textit{et al.} \cite{SSL-STR}            & \multicolumn{2}{c}{72.00} & \multicolumn{2}{c|}{74.80} & \multicolumn{2}{c}{76.20} & \multicolumn{2}{c|}{78.10} & \multicolumn{2}{c}{81.60} & \multicolumn{2}{c|}{84.00} & \multicolumn{2}{c}{52.90} & \multicolumn{2}{c|}{54.70} & \multicolumn{2}{c}{-} & \multicolumn{2}{c|}{-} & \multicolumn{2}{c}{-} & \multicolumn{2}{c|}{-} & \multicolumn{2}{c}{-} & \multicolumn{2}{c}{-} \\
SeqCLR (All-to-instance) \cite{seqCLR}            & \multicolumn{2}{c}{71.03} & \multicolumn{2}{c|}{80.10} & \multicolumn{2}{c}{65.60} & \multicolumn{2}{c|}{74.50} & \multicolumn{2}{c}{75.47} & \multicolumn{2}{c|}{82.56} & \multicolumn{2}{c}{51.37} & \multicolumn{2}{c|}{60.71} & \multicolumn{2}{c}{50.23} & \multicolumn{2}{c|}{63.88} & \multicolumn{2}{c}{42.71} & \multicolumn{2}{c|}{54.86} & \multicolumn{2}{c}{63.06} & \multicolumn{2}{c}{72.39} \\
SeqCLR (Frame-to-instance) \cite{seqCLR}          & \multicolumn{2}{c}{70.37} & \multicolumn{2}{c|}{78.90} & \multicolumn{2}{c}{66.77} & \multicolumn{2}{c|}{75.58} & \multicolumn{2}{c}{74.79} & \multicolumn{2}{c|}{83.25} & \multicolumn{2}{c}{52.67} & \multicolumn{2}{c|}{62.45} & \multicolumn{2}{c}{53.33} & \multicolumn{2}{c|}{64.19} & \multicolumn{2}{c}{43.40} & \multicolumn{2}{c|}{57.29} & \multicolumn{2}{c}{63.41} & \multicolumn{2}{c}{72.69} \\
SeqCLR (Window-to-instance) \cite{seqCLR}         & \multicolumn{2}{c}{71.60} & \multicolumn{2}{c|}{80.37} & \multicolumn{2}{c}{65.38} & \multicolumn{2}{c|}{74.19} & \multicolumn{2}{c}{76.65} & \multicolumn{2}{c|}{82.37} & \multicolumn{2}{c}{51.47} & \multicolumn{2}{c|}{61.34} & \multicolumn{2}{c}{52.87} & \multicolumn{2}{c|}{64.65} & \multicolumn{2}{c}{41.32} & \multicolumn{2}{c|}{58.33} & \multicolumn{2}{c}{63.58} & \multicolumn{2}{c}{72.81} \\  \noalign{\smallskip} \hline \noalign{\smallskip}
\textbf{Ours}                                & \multicolumn{2}{c}{76.70}  & \multicolumn{2}{c|}{81.23} & \multicolumn{2}{c}{72.64} & \multicolumn{2}{c|}{76.66} & \multicolumn{2}{c}{80.59} & \multicolumn{2}{c|}{84.63} & \multicolumn{2}{c}{60.23} & \multicolumn{2}{c|}{62.98} & \multicolumn{2}{c}{61.40} & \multicolumn{2}{c|}{64.81} & \multicolumn{2}{c}{53.82} & \multicolumn{2}{c|}{60.07} & \multicolumn{2}{c}{70.27} & \multicolumn{2}{c}{73.98} \\
\textbf{Ours w/ SeqCLR (All-to-instance)}    & \multicolumn{2}{c}{78.73} & \multicolumn{2}{c|}{\textbf{83.47}} & \multicolumn{2}{c}{75.58} & \multicolumn{2}{c|}{79.60} & \multicolumn{2}{c}{83.65} & \multicolumn{2}{c|}{87.19} & \multicolumn{2}{c}{64.37} & \multicolumn{2}{c|}{\textbf{68.27}} & \multicolumn{2}{c}{66.36} & \multicolumn{2}{c|}{68.22} & \multicolumn{2}{c}{55.90} & \multicolumn{2}{c|}{64.24} & \multicolumn{2}{c}{72.99} & \multicolumn{2}{c}{77.13} \\
\textbf{Ours w/ SeqCLR (Frame-to-instance)}  & \multicolumn{2}{c}{\textbf{80.50}} & \multicolumn{2}{c|}{83.37} & \multicolumn{2}{c}{\textbf{76.20}} & \multicolumn{2}{c|}{79.44} & \multicolumn{2}{c}{84.34} & \multicolumn{2}{c|}{\textbf{86.90}} & \multicolumn{2}{c}{\textbf{65.34}} & \multicolumn{2}{c|}{67.45} & \multicolumn{2}{c}{\textbf{67.75}} & \multicolumn{2}{c|}{\textbf{70.08}} & \multicolumn{2}{c}{\textbf{58.33}} & \multicolumn{2}{c|}{\textbf{65.97}} & \multicolumn{2}{c}{\textbf{74.56}} & \multicolumn{2}{c}{\textbf{77.36}} \\
\textbf{Ours w/ SeqCLR (Window-to-instance)} & \multicolumn{2}{c}{79.87} & \multicolumn{2}{c|}{83.23} & \multicolumn{2}{c}{74.34} & \multicolumn{2}{c|}{\textbf{80.53}} & \multicolumn{2}{c}{\textbf{85.22}} & \multicolumn{2}{c|}{86.70} & \multicolumn{2}{c}{63.70} & \multicolumn{2}{c|}{67.55} & \multicolumn{2}{c}{66.36} & \multicolumn{2}{c|}{69.30} & \multicolumn{2}{c}{57.99} & \multicolumn{2}{c|}{64.93} & \multicolumn{2}{c}{73.74} & \multicolumn{2}{c}{77.19} \\  \noalign{\smallskip} \hline
\end{tabular}
}
\end{center}
\end{table*}
\setlength{\tabcolsep}{1.4pt}

Similarly, in Table \ref{table:STR}, we compare our method's performance on scene-text recognition task on word prediction accuracy following the same training principles used to train the handwriting recognition models. The last column in Table \ref{table:STR} represents the combined total accuracy \textit{i.e.} all the scene-text datasets combined together for evaluation. The results corroborate the superiority of our methods in the semi-supervised scene-text recognition scenario, with an average gain in the total accuracy of $\approx 6.7\%$ and $\approx 1.2\%$ for model trained with $5\%$ and $10\%$ of the total labeled data on the vanilla PL-SSL strategy, respectively.

\begin{figure}[ht]
    \centering
    \includegraphics[width=\linewidth]{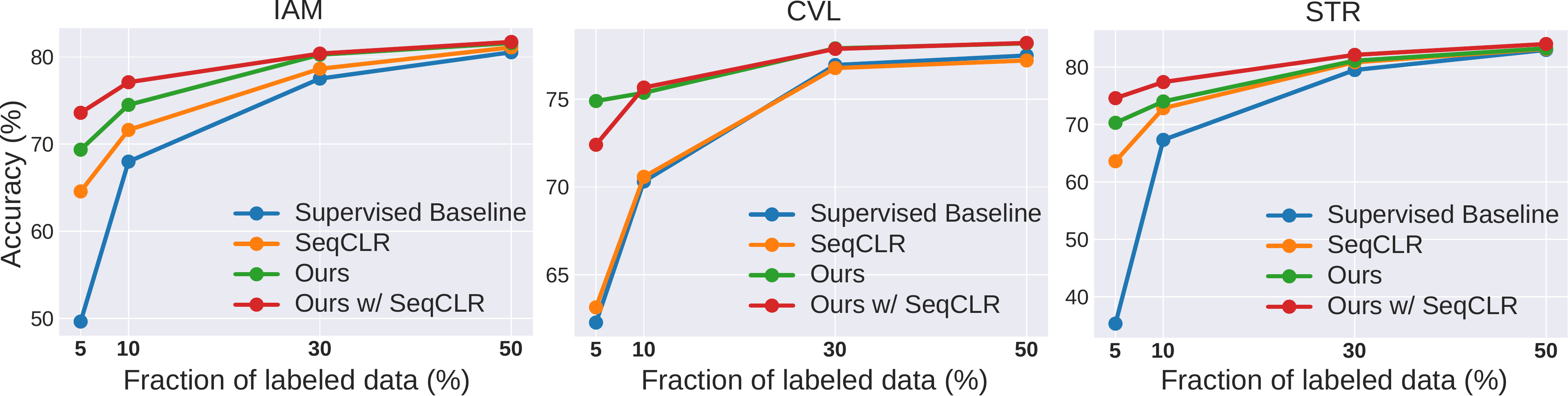}
    \caption{Word prediction accuracy trends in the semi-supervised setting with 5\%, 10\%, 30\%, and 50\% of the total available labeled data used as $\mathcal{D}_{l}$ , for the IAM, CVL, and the combined scene-text recognition (STR) datasets. }\label{fig:accuracy_comparison}
\end{figure}

Moreover, in Figure \ref{fig:accuracy_comparison} we visualize the word prediction accuracy on the handwriting datasets (IAM and CVL) and the scene-text recognition (STR) datasets (all datasets combined) at different fractions of the labels used on the supervised baseline, SeqCLR, and our proposed method with and without SeqCLR pre-training. Note that the SeqCLR trends are shown for its best performing variant.

\subsection{Ablation Studies}
\begin{figure}
    \centering
    \includegraphics[width=\linewidth]{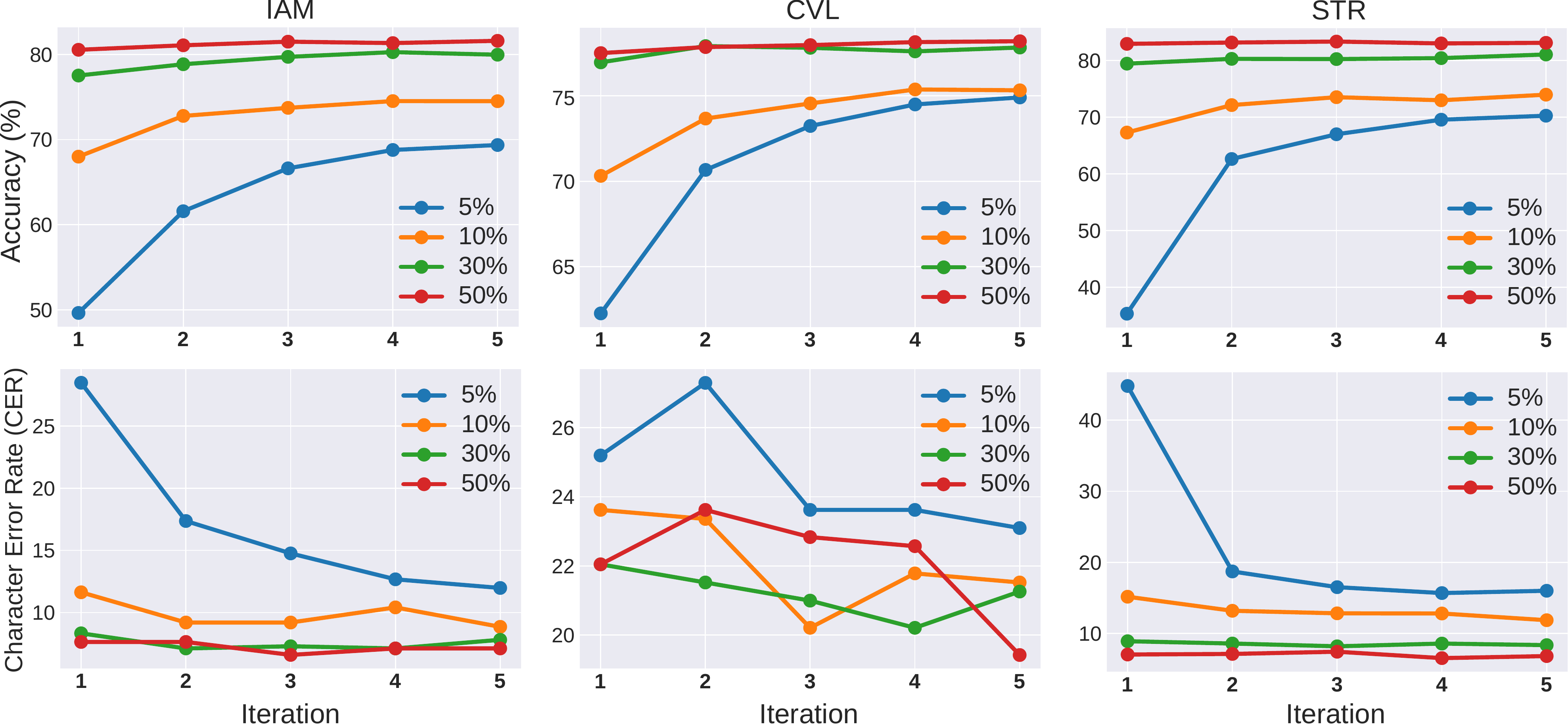}
    \caption{Iteration-wise metric evolution of the pseudo-labeling based semi-supervised learning methodology with randomly initialized weights.}
    \label{fig:metric_trends}
\end{figure}

\begin{figure}
    \centering
    \begin{subfigure}[b]{0.48\linewidth}
    \centering
    \includegraphics[width=\linewidth]{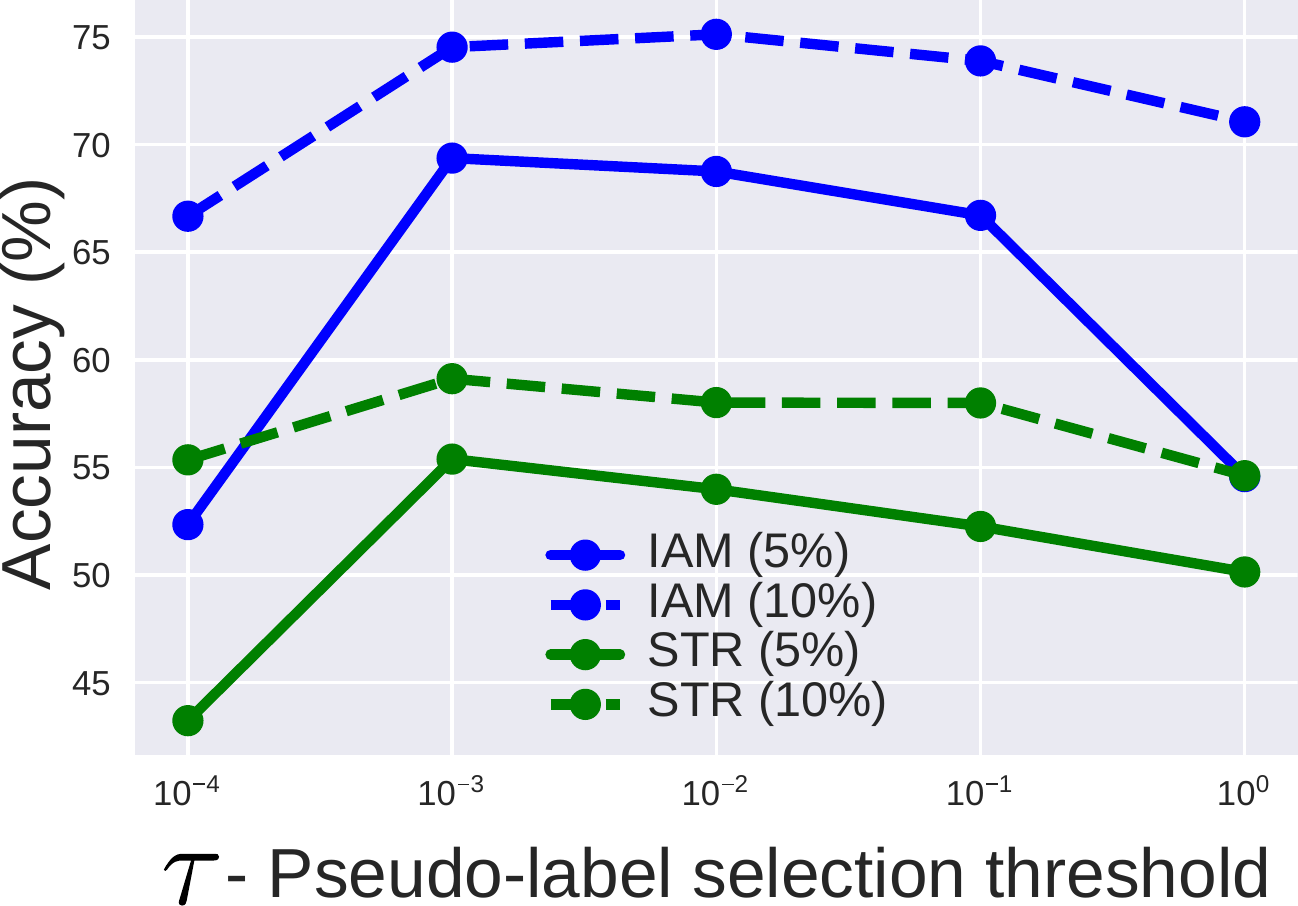}
    \caption{ }
    \label{fig:tau}
    \end{subfigure}
    \hfill
    \begin{subfigure}[b]{0.48\linewidth}
    \centering
    \includegraphics[width=\linewidth]{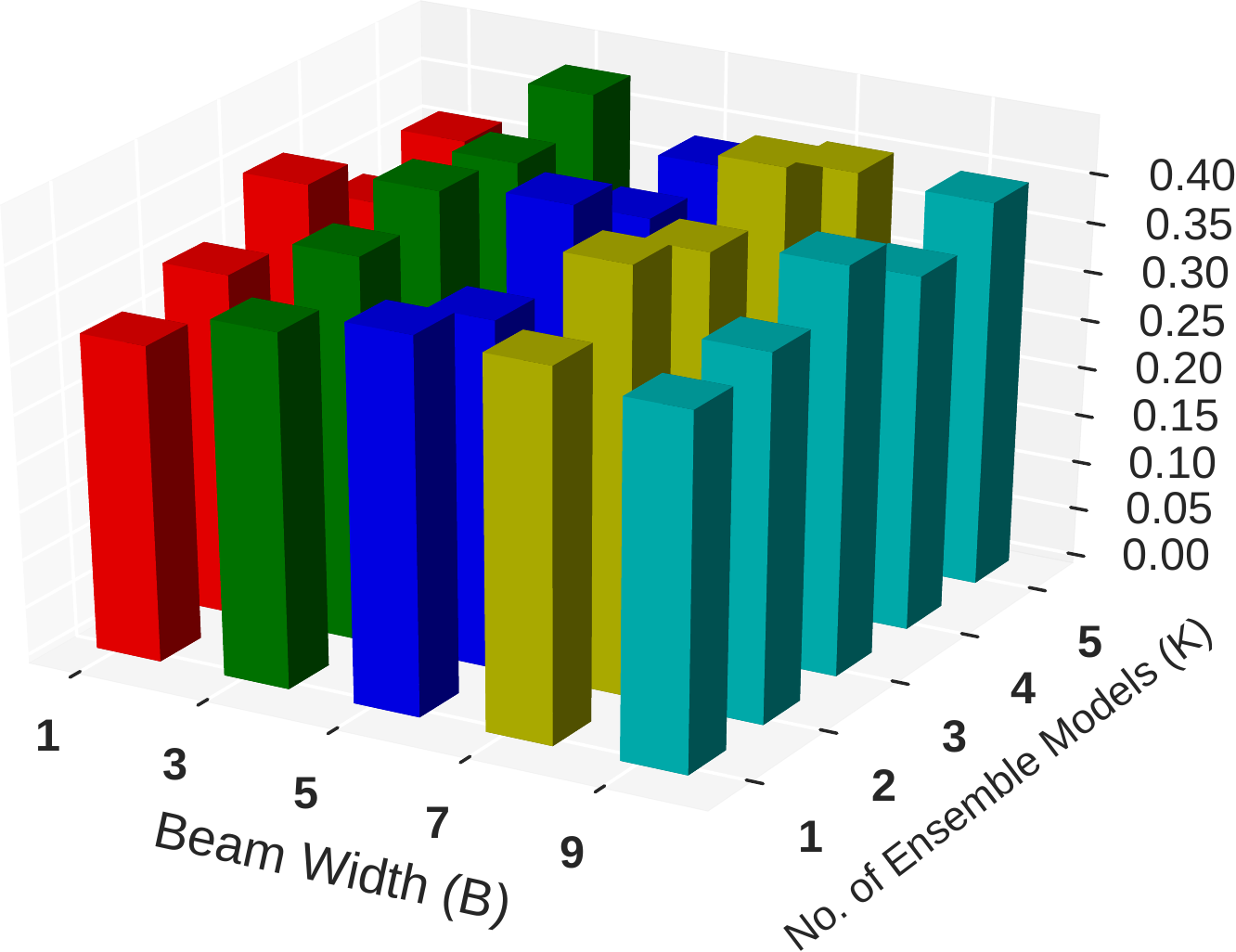}
    \caption{ }
    \label{fig:PRR_ablation}
    \end{subfigure}

    \caption{(a) Impact of $\tau$ at different values, on IAM and STR validation accuracy with 5\% (solid lines) and 10\% (dashed lines) of labeled data. (b) \textit{Prediction rejection ratio} (\textit{PRR}) variation of the employed uncertainty estimation across different values of  beam-width ($B$) and the no. of ensembles ($K$), on the model trained on IAM dataset, based on the word prediction error metric. Note that values corresponding to the the no. of ensembles $K=1$ depicts \emph{deterministic inference}.}
    \label{fig:PRR_tau}
\end{figure}

\paragraph{On iteration-wise performance evolution and the $\tau$ hyperparameter: }Since the introduced methodology is iterative, and consists of  multiple self-training steps, in Figure \ref{fig:metric_trends} we visualize the vanilla  PL-SSL method's performance on word prediction accuracy and CER at the end of each supervised training iteration, with different portions of labeled training dataset used to train the seed model, for the IAM, CVL, and combined STR datasets. We observe improvement in the performance in the initial iteration and then a saturation in the performance, this can be attributed to scenario when the set threshold hyperparameter $\tau$ does not allow new samples to be included for the further rounds, thus suggesting a constant $\tau$ parameter achieves a state of equilibrium and is not able to include new samples, which motivates future research directions to learn an iteration-wise adaptive threshold parameter or a curriculum of threshold which could break the equilibrium by having variable $\tau$ values, for the inclusion of additional samples at later iterations.

\begin{figure}[t]
    \centering
    \includegraphics[width=.55\linewidth]{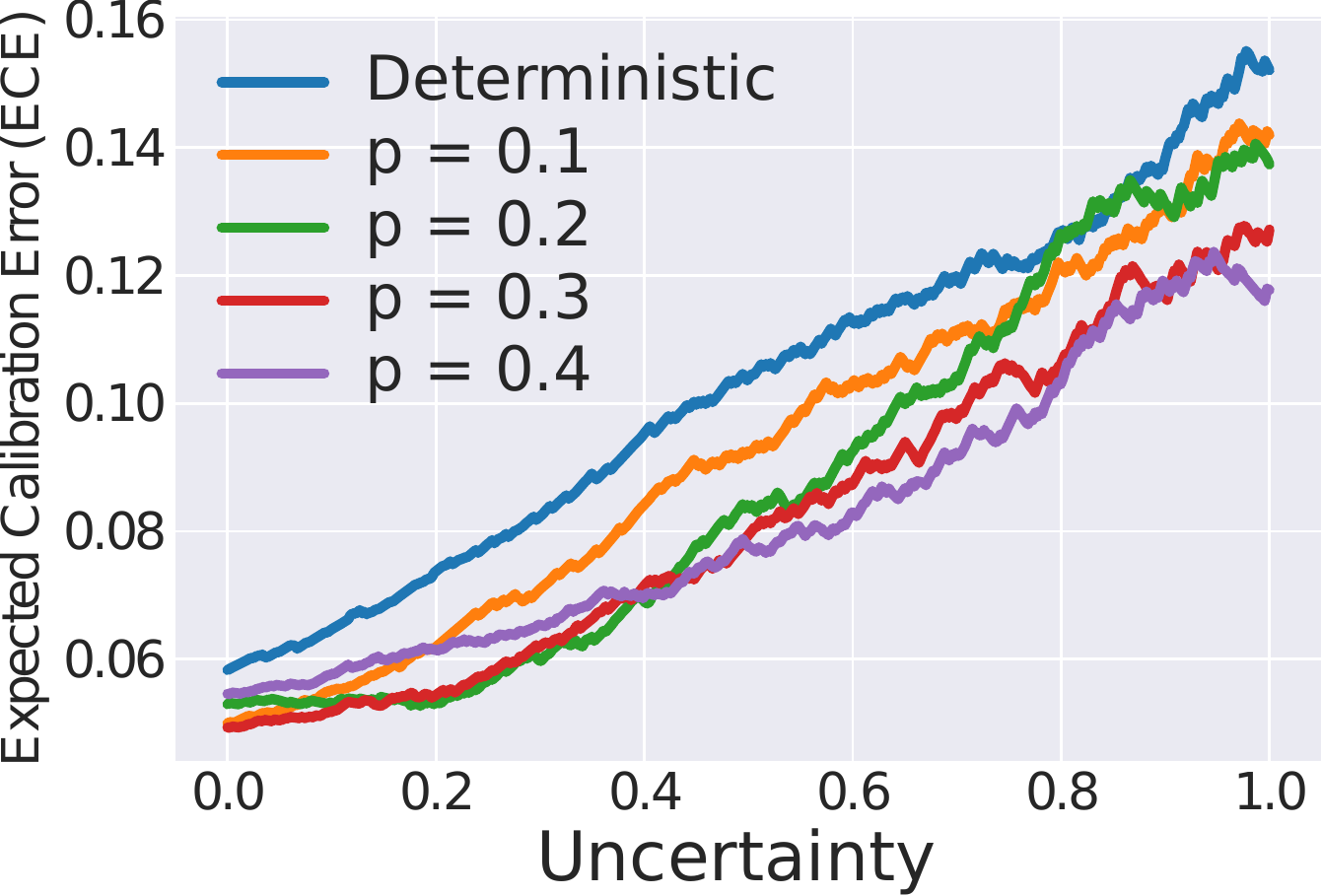}
    \caption{The link between prediction uncertainty and expected calibration error (ECE) is depicted here. With different dropout probability $p$, the ECE of the selected subset diminishes as the uncertainty of the picked pseudo-labels decreases. Note that, compared to \emph{deterministic-inference}, the MC-Dropout-based model ensembling provides a lower calibration error for a fixed number of elements in the uncertainty-based selected subset. }
    \label{fig:dropout_ece}
\end{figure}

As established above, the threshold hyperparameter $\tau$ plays a critical role in the eﬀectiveness of the proposed method, to understand its effect on the performance on PL-SSL, we run experiments on the IAM, CVL and STR datasets with different $\tau$ values in $\{10^{-i}\}_{i=0}^{4}$ and note the best performance among the self-training iterations, on their respective validation set. In Figure \ref{fig:tau}, we visualize the validation performance at different $\tau$ values and observe that when $\tau$ is significantly low, the selected examples are less informative, since they are too low in number for improving the model's performance, especially if large proportion of labeled data is used. If too high, the selected samples become too noisy for training. For all our experiments, we use a fixed $\tau$ value of $10^{-2}$.
\paragraph{On optimal Beam-width ($B$) and No. of ensemble models ($K$) combination:} To compute the uncertainty measure defined in (\ref{eq:6}), we require a finite set of highly probable predictions to approximate the hypotheses space and the predictive distribution of each hypothesis is approximated as described in expression (\ref{eq:1}) by taking the expectation of the virtually generated ensemble. To identify the optimum beam-width ($B$) and the no. of ensemble models ($K$) combination, we visualize the variation of the \textit{PRR}, described in subsection \ref{rue}, with respect to a finite set of values of $B$ and $K$ as shown in Figure \ref{fig:PRR_ablation}. From the plot we observe no specific trend on the combination of $B$ and $K$ which can generalize the hyperparameter selection, therefore, in all our experiments, we used a fixed value for $B$ and $K$ equal to 5, in practice, one can use the validation set to find the optimal combination by observing the \textit{PRR}.
\paragraph{Effect of dropout on the prediction distribution:}
To corroborate our motivation, we conduct experiments to observe the effect of dropout when computing the predictive distribution, we evaluate the quality of the predictive distribution by observing the Expected Calibration Error (ECE) at distinct dropout probability values. In Table \ref{table:dropout}, we present the ECE values for multiple datasets, at different dropout probability ($p$) values. As we increase the dropout value, the ECE reduces, hence, demonstrating the utility of dropout in computing the predictive distribution which is eventually used to compute the \textit{total-uncertainty}. Finally, in Figure \ref{fig:dropout_ece}, we show the relationship between the prediction uncertainty and the calibration error (ECE). When pseudo-labels with more certain predictions are chosen, the calibration error for that subset of pseudo-labels is considerably reduced. As a result, the predictions are more likely to result in an accurate pseudo-label for that subset of labels. 


\setlength{\tabcolsep}{4pt}
\begin{table}[t]
\begin{center}
\resizebox{0.7\linewidth}{!}{
\begin{tabular}{l|ccccc}
\hline\noalign{\smallskip}
\multirow{2}{*}{\textbf{Dataset}} & \multicolumn{5}{c}{\textbf{Dropout Probability ($p$)}} \\
                         & \textbf{0.0}    & \textbf{0.1}     & \textbf{0.2}    & \textbf{0.3}    & \textbf{0.4}    \\ \noalign{\smallskip}\hline\noalign{\smallskip}
IAM                      & 0.1551 & 0.1475 & 0.1409 & 0.1335 & 0.1237 \\
CVL                      & 0.1914 & 0.1819  & 0.1718   & 0.1601   & 0.1513 \\
STR                      & 0.4381 & 0.4266  & 0.4169 & 0.3939 & 0.3759 \\ \noalign{\smallskip}\hline
\end{tabular}
}
\end{center}
\caption{Expected Calibration Error (ECE) across different dropout probability values ($p$) with $K=5$ ensembles, as used in all the experiments, trained with all the samples. Note that values corresponding to $p=0.0$ denotes \emph{deterministic inference}.}\label{table:dropout}
\end{table}
\section{Discussion}
\label{sec:conclusion}
\noindent\textbf{Limitations: }Despite the superior SSL performance, the method discussed is not devoid of certain limitations. Thus, we identify two primary limitations as follows. Firstly, the current framework assumes the same character space for the labeled and the unlabeled examples. Therefore, it may be hard to employ such an SSL framework directly in scenarios when the labeled and the unlabeled examples have character space discrepancy, \textit{e.g.}, cross-lingual text recognition, when labeled examples come from one language and unlabeled samples from the other. The second assumption is based on the presence of dropout layers in the backbone network. In Table \ref{table:dropout}, we observe improvements in the Expected Calibration Error (ECE) due to the presence of dropout. However, the observation made in \cite{UPS} and verified as shown in Figure \ref{fig:dropout_ece}  suggests that selecting pseudo-labeled samples with low uncertainty also helps reduce the calibration error. Additionally, directly estimating uncertainty by \emph{deterministic inference} can also be considered, though it may require having a stricter (lower) sub-set selection threshold $\tau$ to attenuate the calibration error, however the number of pseudo-labeled examples may be low.\\
\textbf{Conclusion: }This paper proposes a pseudo-label generation and an uncertainty-based data selection framework for semi-supervised image-based text recognition. The method uses Beam-Search inference to yield highly probable hypotheses for pseudo-label assignment of the unlabelled samples. Furthermore, it adopts an ensemble of models obtained by MC-Dropout to get a robust estimate of the uncertainty related to the input image that considers both the character-level and word-level predictive distributions. Moreover, we show superior SSL performance in handwriting and scene-text recognition tasks. In addition, we also utilize prediction rejection curves to demonstrate the correlation of the uncertainty estimate with the word prediction error rate.

\noindent\textbf{Acknowledgement:} Work partially supported by HP and \href{https://www.nsf.gov/awardsearch/showAward?AWD_ID=1737744&HistoricalAwards=false}{National Science Foundation (1737744)}.

\input{appendix.tex}


{\small
\bibliographystyle{ieee_fullname}
\bibliography{egbib}
}

\end{document}

%% file: appendix.tex
\pdfoutput=1
\appendix
\label{sec:appendix}

\setlength{\tabcolsep}{4pt}
\begin{table}[ht]
\centering
\caption{ResNet architecture configuration for the text recognition model. Here, $c$, $k$, $s$, and $p$ stand for no. of channels, filter size, stride, and padding, respectively.}\label{table_architecture}

\begin{tabular}{|c|cc|c|}
\hline
\textbf{Layers}               & \multicolumn{2}{c|}{\textbf{Configurations}} & \textbf{Output}                \\ \hline
Input                         & \multicolumn{2}{c|}{grayscale image}         & $100 \times 32$                \\ \hline
Conv1                         & $c: 32$               & $k: 3 \times 3$      & $100 \times 32$                \\ \hline
Conv2                         & $c: 64$               & $k: 3 \times 3$      & $100 \times 32$                \\ \hline
\multicolumn{1}{|l|}{Dropout} & \multicolumn{2}{c|}{-}                       & $100 \times 32$                \\ \hline
Pool1                         & $k: 2 \times 2$       & $s: 2 \times 2$      & $50 \times 16$                 \\ \hline
Block1                        & \multicolumn{2}{c|}{$\begin{bmatrix}
c: 128, \ & \ k: 3\times3\\
c: 128, \ & \ k: 3\times3
\end{bmatrix} \times 1$}             & $50 \times 16$                 \\ \hline
Conv3                         & $c: 128$              & $k: 3 \times 3$      & $50 \times 16$                 \\ \hline
\multicolumn{1}{|l|}{Dropout} & \multicolumn{2}{c|}{-}                       & $50 \times 16$                 \\ \hline
Pool2                         & $k: 2 \times 2$       & $s: 2 \times 2$      & $25 \times 8$                  \\ \hline
Block2                        & \multicolumn{2}{c|}{$\begin{bmatrix}
c: 256, \ & \ k: 3\times3\\
c: 256, \ &  \ k: 3\times3
\end{bmatrix}\times 2$}             & $25 \times 8$                  \\ \hline
Conv4                         & $c: 256$              & $k: 3 \times 3$      & $25 \times 8$                  \\ \hline
\multicolumn{1}{|l|}{Dropout} & \multicolumn{2}{c|}{-}                       & $25 \times 8$                  \\ \hline
\multirow{2}{*}{Pool3}        & $k: 2 \times 2$       &                      & \multirow{2}{*}{$26 \times 4$} \\
                              & $s: 1 \times 2$       & $p: 1 \times 0$      &                                \\ \hline
Block3                        & \multicolumn{2}{c|}{$\begin{bmatrix}
c: 512, \ &  \ k: 3\times3\\
c: 256, \ &  \ k: 3\times3
\end{bmatrix}\times 5$}             & $26 \times 4$                  \\ \hline
Conv5                         & $c: 512$              & $k: 3 \times 3$      & $26 \times 4$                  \\ \hline
\multicolumn{1}{|l|}{Dropout} & \multicolumn{2}{c|}{-}                       & $26 \times 4$                  \\ \hline
Block4                        & \multicolumn{2}{c|}{$\begin{bmatrix}
c: 512, \ &  \ k: 3\times3\\
c: 512, \ &  \ k: 3\times3
\end{bmatrix}\times 3$}             & $26 \times 4$                  \\ \hline
\multirow{2}{*}{Conv6}        & $c: 512$              & $k: 2 \times 2$      & \multirow{2}{*}{$27 \times 2$} \\
                              & $s: 1 \times 2$       & $p: 1 \times 0$      &                                \\ \hline
\multirow{2}{*}{Conv7}        & $c: 512$              & $k: 2 \times 2$      & \multirow{2}{*}{$26 \times 1$} \\
                              & $s: 1 \times 2$       & $p: 0 \times 0$      &                                \\ \hline
\multicolumn{1}{|l|}{Dropout} & \multicolumn{2}{c|}{-}                       & $26 \times 1$                  \\ \hline
\end{tabular}
\end{table}

\section{Dataset Descriptions}
\subsection{Handwriting Recognition Datasets}
\paragraph{CVL \cite{CVL}:}
310 individual writers contributed to this handwritten English text dataset, which was divided into two parts: training and testing. 27 of the writers created 7 texts, while the remaining 283 created 5 texts. 

\paragraph{IAM \cite{IAM}:}
657 different writers contributed to this English handwritten text dataset, which was partitioned into writer independent training, validation, and test.

\subsection{Scene-Text Datasets}
\paragraph{ICDAR-15 (IC15) \cite{ICDAR15}:}
The images in the dataset were gathered by people wearing Google Glass, therefore many of the images have perspective inscriptions and some are fuzzy.
It includes 4,468 training images and 2,077 evaluation images. 

\paragraph{ICDAR-13 {IC13} \cite{ICDAR13}:}
The dataset was created for the ICDAR 2013 Robust Reading competition. It contains 848 images for training and 1,015 images for evaluation.

\paragraph{IIIT5k-Words (IIIT) \cite{IIIT}:}
Google image searches with query phrases like "billboards" and "movie posters" yielded the text-images. It includes 2,000 training photos and 3,000 evaluation images. 

\paragraph{Street View Text (SVT) \cite{SVT}:}
The dataset is prepared based on Google Street View and includes text included in street photos. It includes 257 training images and 647 evaluation images. 

\paragraph{SVT Perspective (SVTP) \cite{SVTP}:}
Similar to SVT, SVTP is gathered from Google Street View. In contrast to SVT, SVTP features a large number of perspective texts. It includes 645 images for evaluation. 

\paragraph{CUTE80 (CUTE) \cite{CUTE}:}
CUTE contains curved text images. The images are captured by a digital camera or collected from the Internet. It contains 288 cropped images for evaluation.

\paragraph{COCO-Text (COCO) \cite{veit2016coco}:}
This dataset is created from text instances from the original MS-COCO dataset \cite{COCO}.

\paragraph{RCTW \cite{RCTW}:}
RCTW stands for the \textbf{R}eading \textbf{C}hinese \textbf{T}ext in the \textbf{W}ild dataset. Primarily containing Chinese text. Nonetheless, we used the non-Chinese text images in the training set.

\paragraph{Uber-Text (Uber) \cite{UberText}:}
Bing Maps Streetside was used to obtain Uber-Text image data. Many of them are house numbers, while others are text on billboards. 

\paragraph{Arbritary-shaped Text (ArT) \cite{Art}:}
This dataset contains images with perspective, rotation, or curved text.

\paragraph{Large-scale Street View Text (LSVT) \cite{LSVT1,LSVT2}:}
Data collected from the streets in China. Thus, most of the text is in Chinese.

\paragraph{Multi-Lingual Text (MLT) \cite{MLT19}:}
This dataset is created to recognize multi-lingual text. It consists of text images from seven languages:  Arabic, Latin, Chinese, Japanese, Korean, Bangla, and Hindi.

\paragraph{Reading Chinese Text on Signboard (ReCTS) \cite{ReCTS}:}
Created for the Reading Chinese Text on Signboard competition. It features a large number of irregular texts that are grouped in various layouts or written in different typefaces.

\paragraph{}
For further extensive details on the used scene-text datasets and the adopted preprocessing we refer the readers to \cite{deeptext,RealSTR_data}.

\section{Text-Recongnition Model Architecture}
We adopt the best performing recognition model used in \cite{seqCLR}, \cite{deeptext}, and \cite{RealSTR_data}, dubbed as \textbf{TRBA} which consists of a thin-plate-spline \cite{STN} \textbf{T}ransformation module, a \textbf{R}esNet-based feature extraction network as used in \cite{deepSTR17}, two \textbf{B}iLSTM layers with 256 hidden units per layer to converts visual features to contextual sequence of features, and lastly an \textbf{A}ttention based LSTM sequential decoder with the hidden state dimension of 256 to convert the sequential features to machine-readable text.  Additionally, in the ResNet backbone we introduce dropout layers for Monte-Carlo sampling as depicted in Table \ref{table_architecture}\footnote{Implementation based on: \url{https://github.com/clovaai/deep-text-recognition-benchmark}}.

\begin{figure*}[ht]
    \centering
    \includegraphics[width=0.24\textwidth]{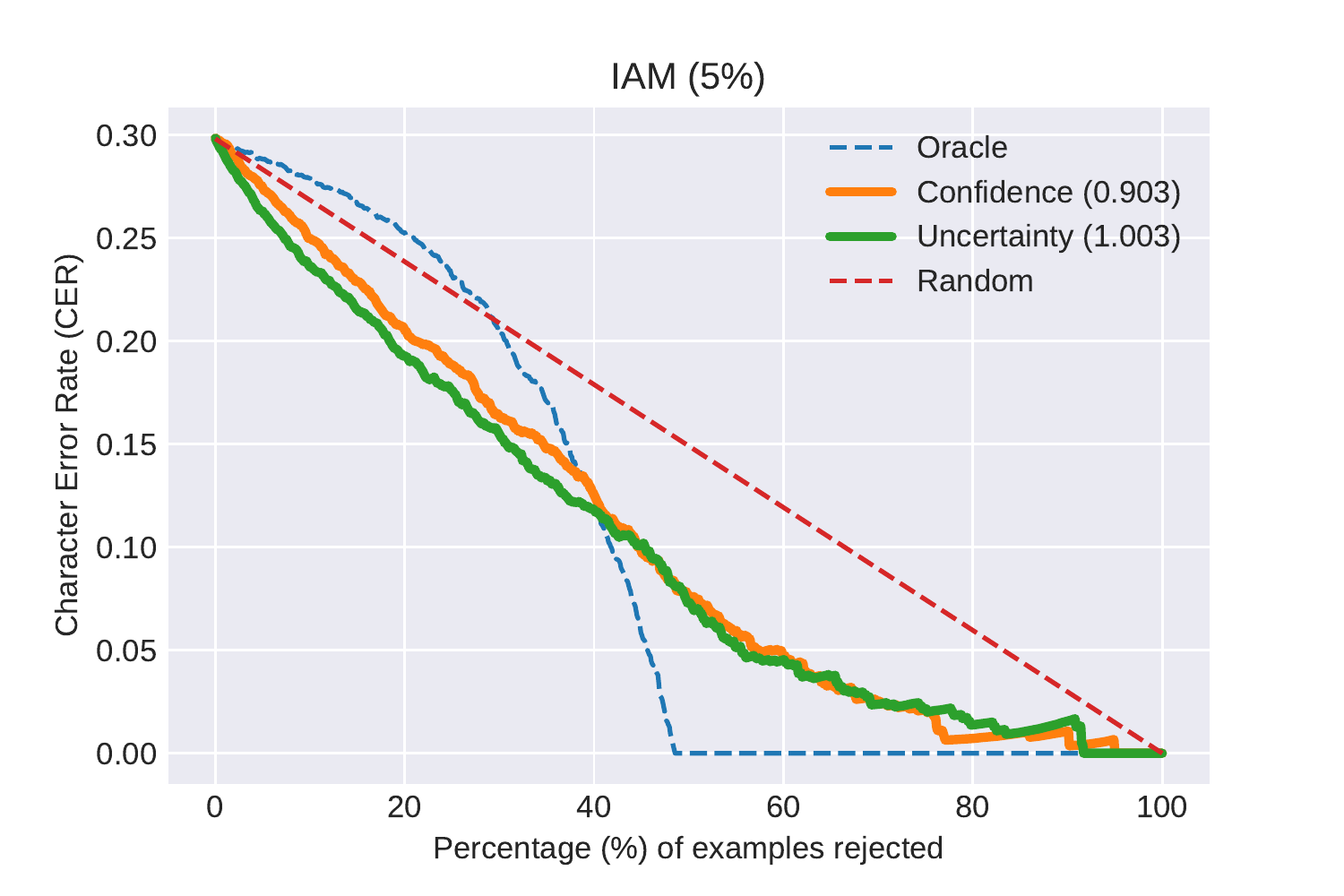}\hfill
    \includegraphics[width=0.24\textwidth]{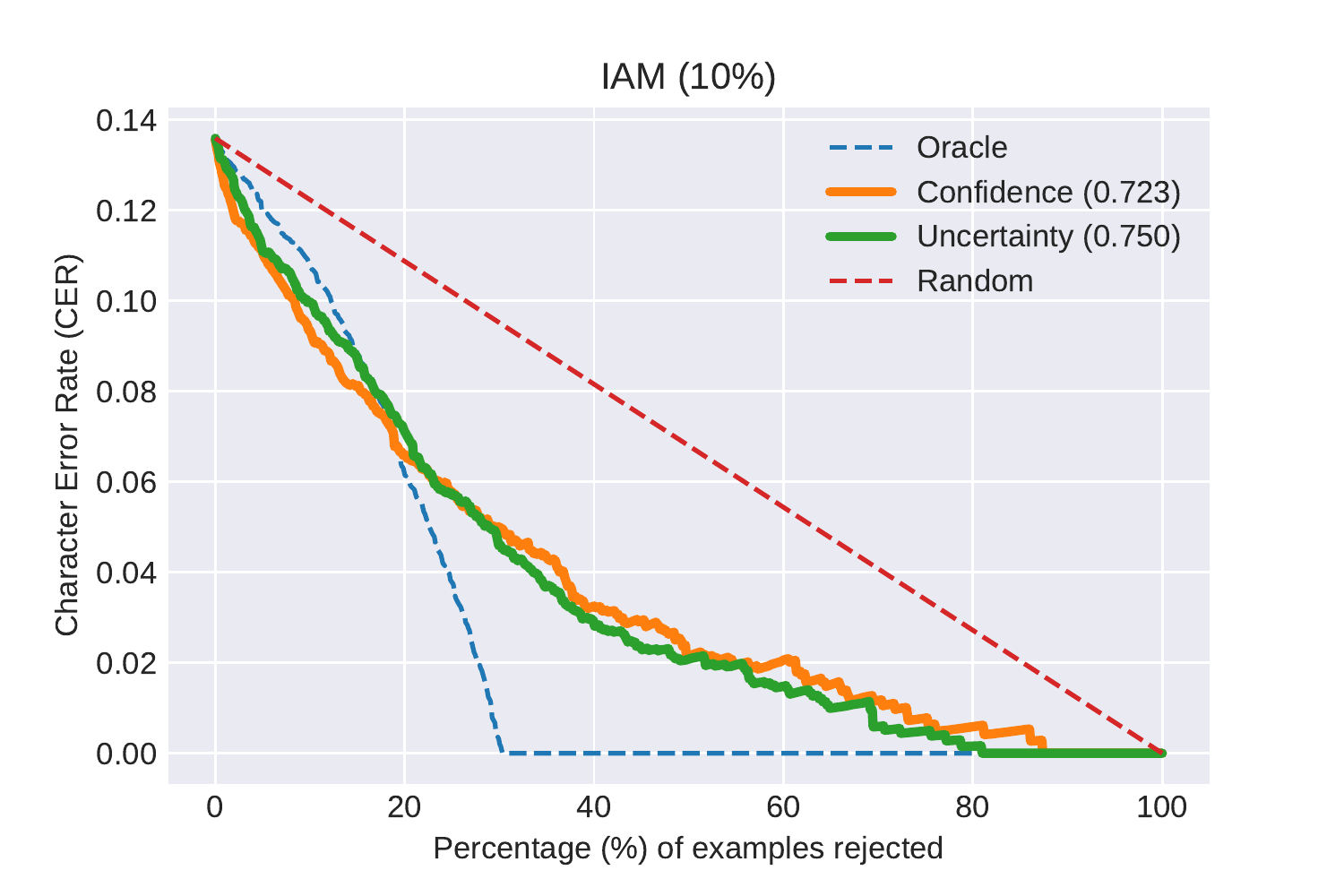}\hfill
    \includegraphics[width=0.24\textwidth]{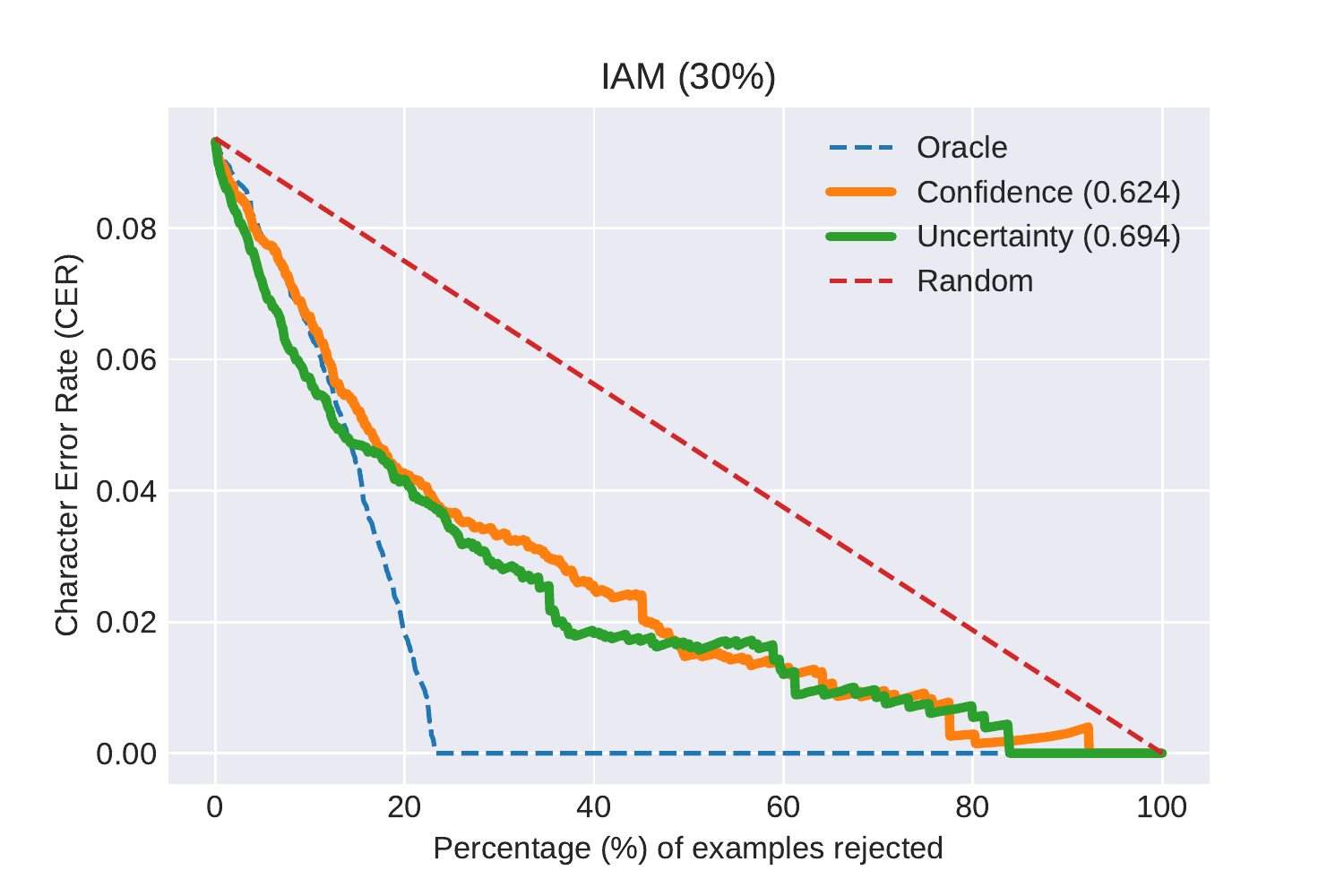}\hfill
    \includegraphics[width=0.24\textwidth]{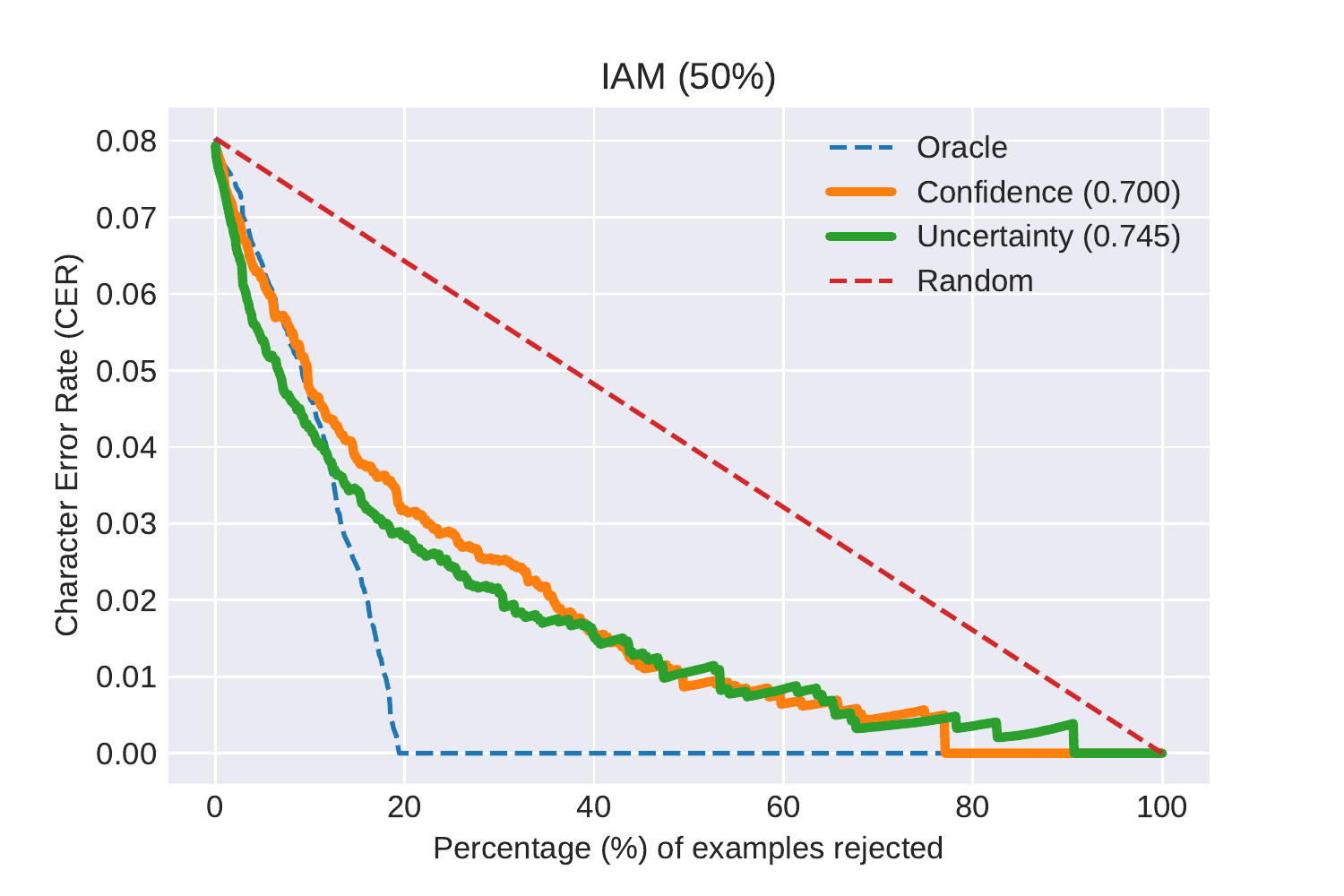}
    \includegraphics[width=0.24\textwidth]{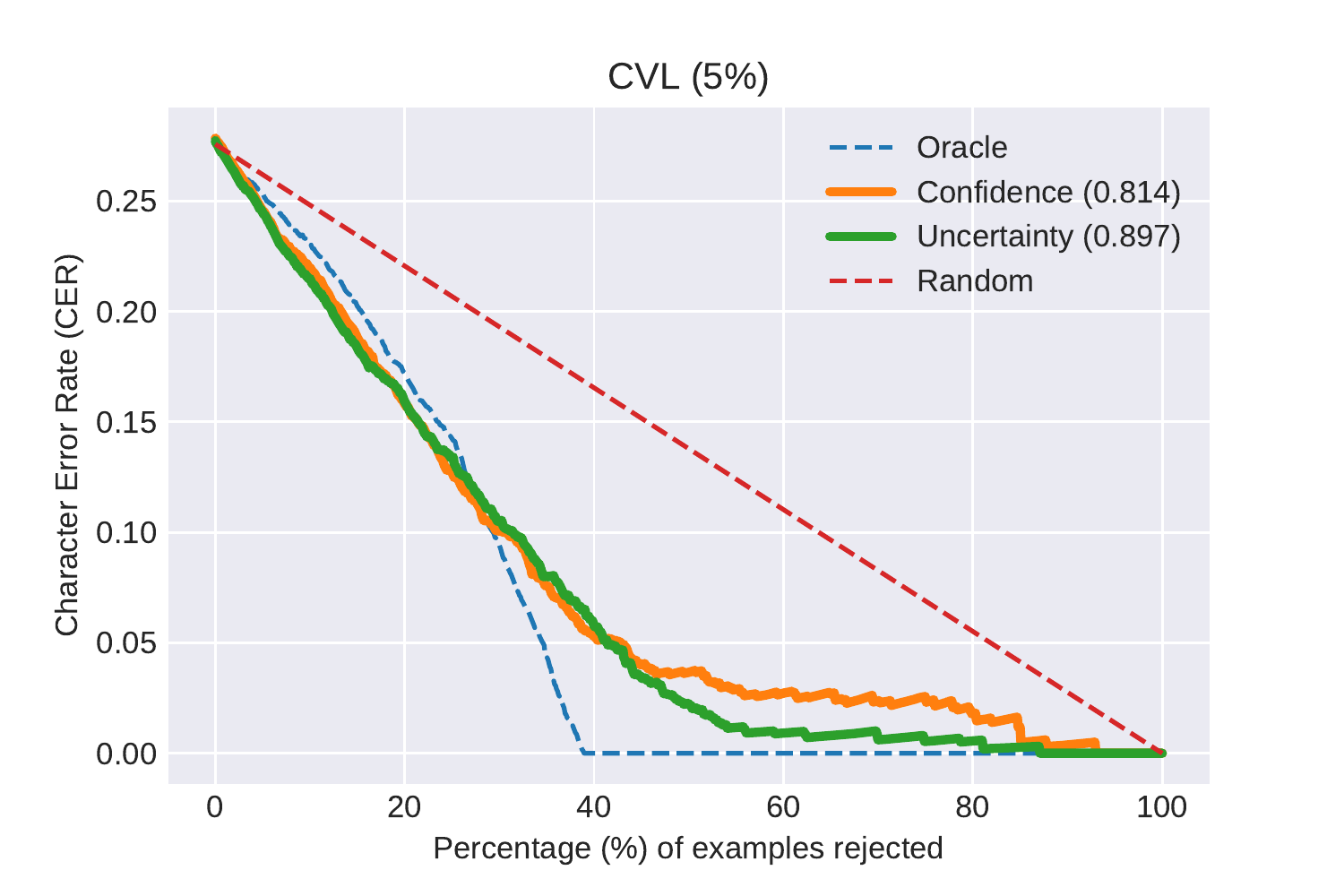}\hfill
    \includegraphics[width=0.24\textwidth]{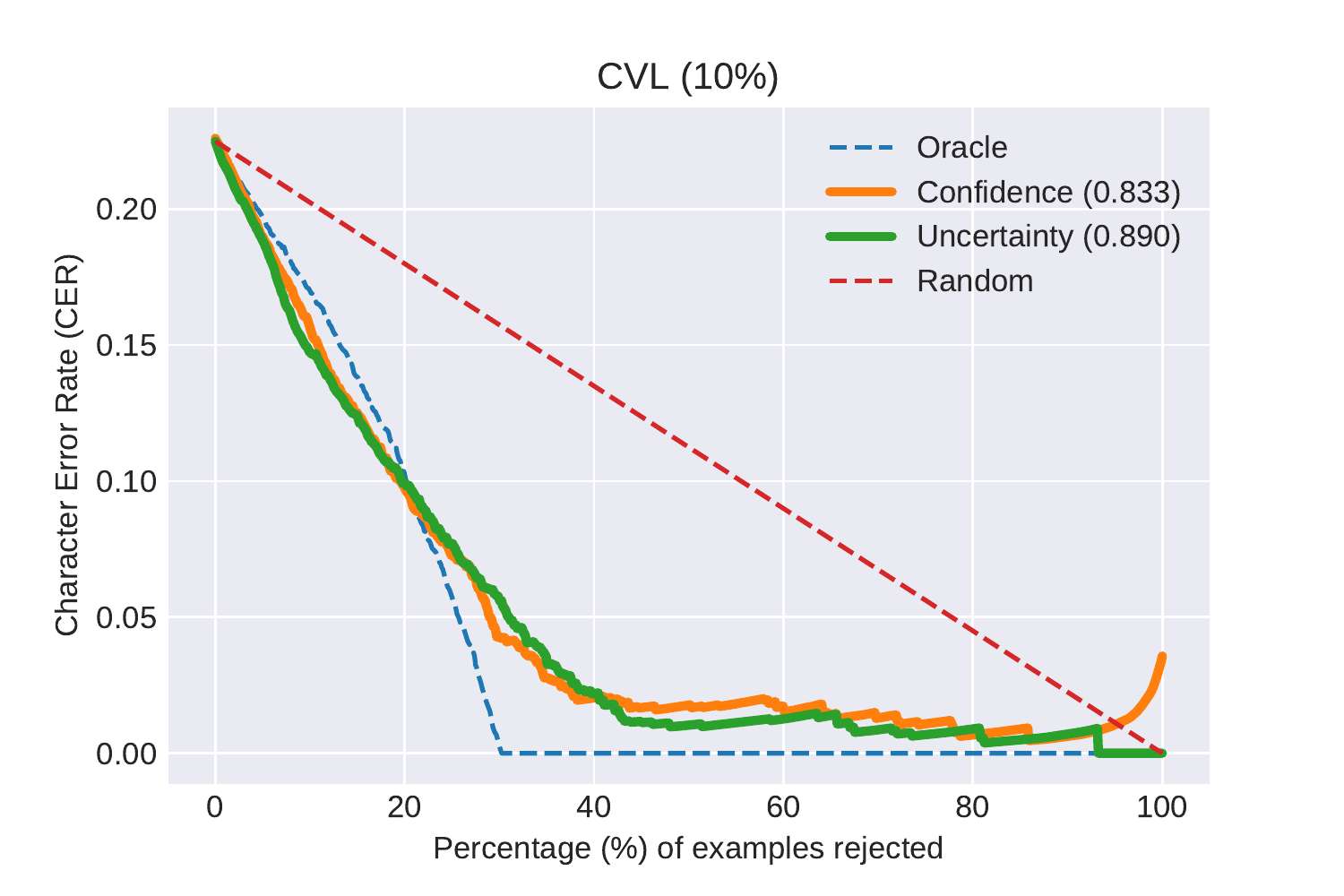}\hfill
    \includegraphics[width=0.24\textwidth]{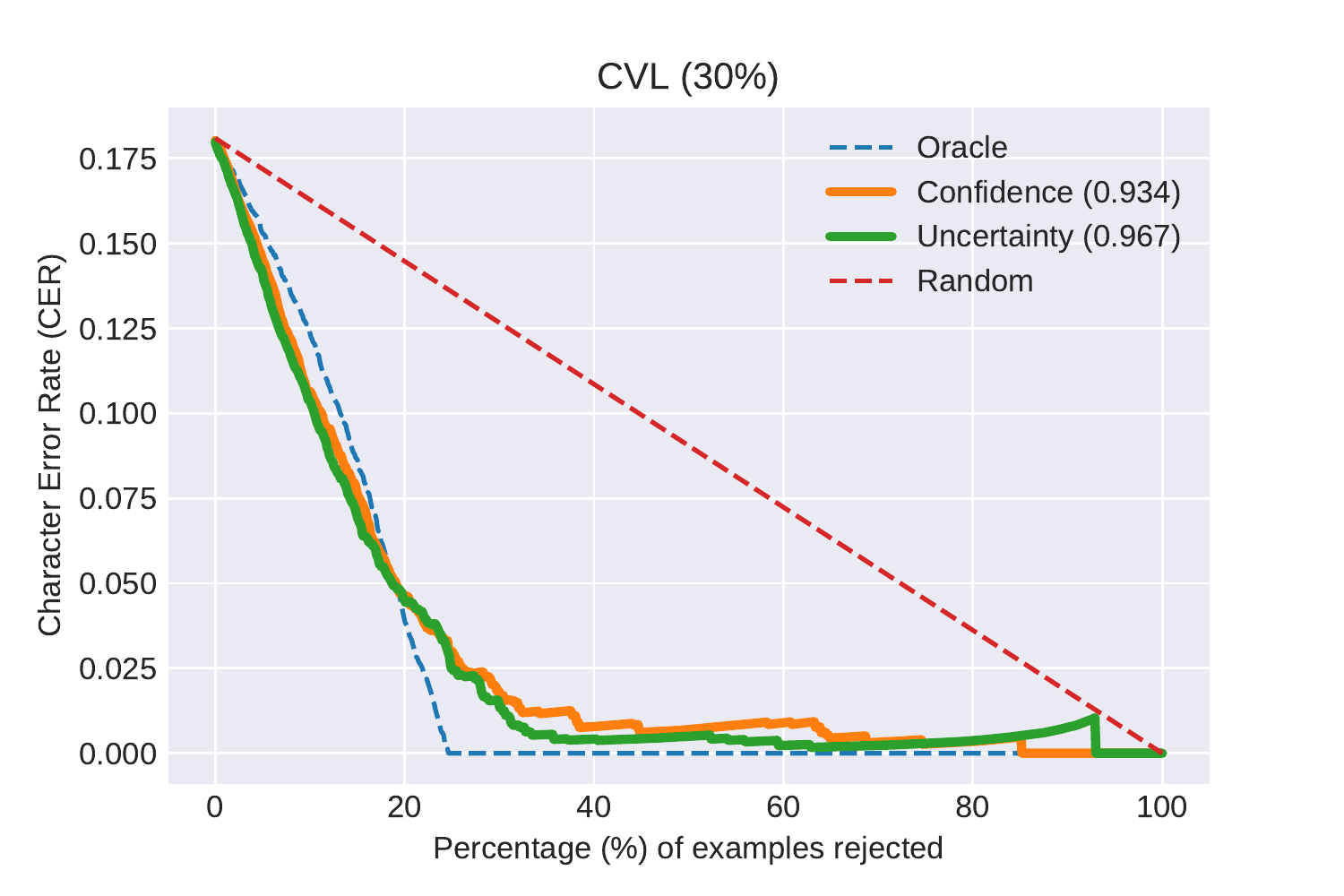}\hfill
    \includegraphics[width=0.24\textwidth]{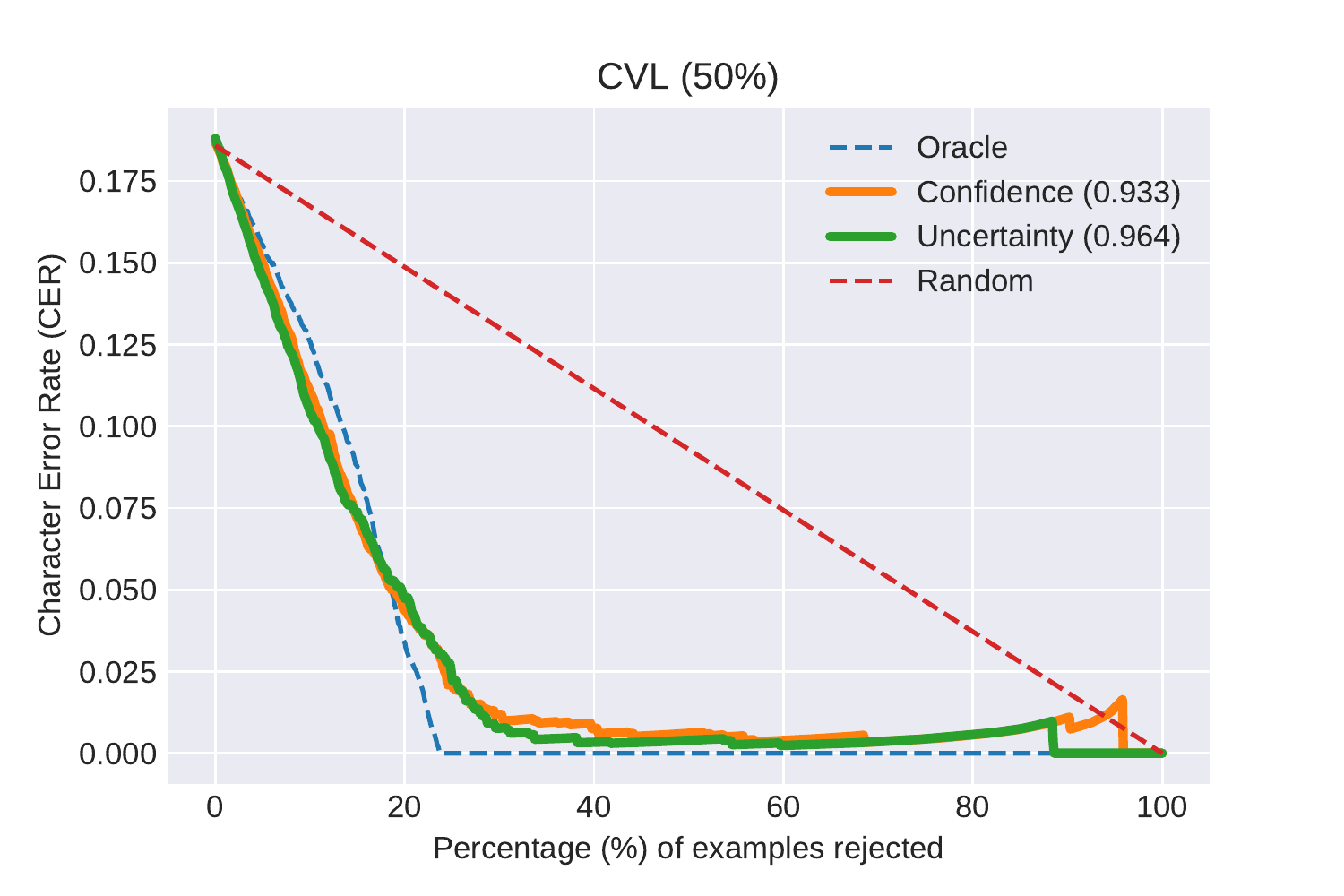}
    \\[\smallskipamount]
    \includegraphics[width=0.24\textwidth]{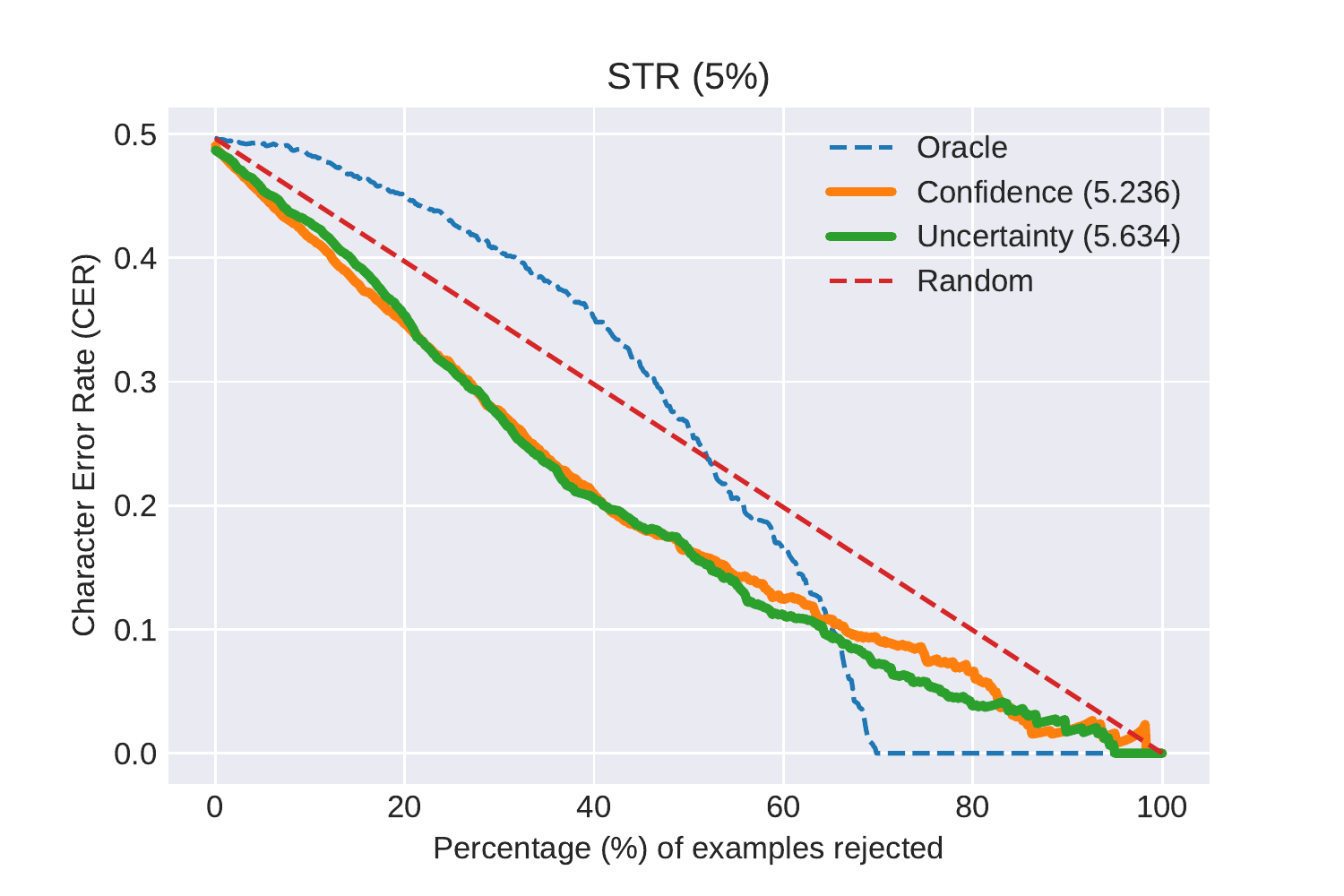}\hfill
    \includegraphics[width=0.24\textwidth]{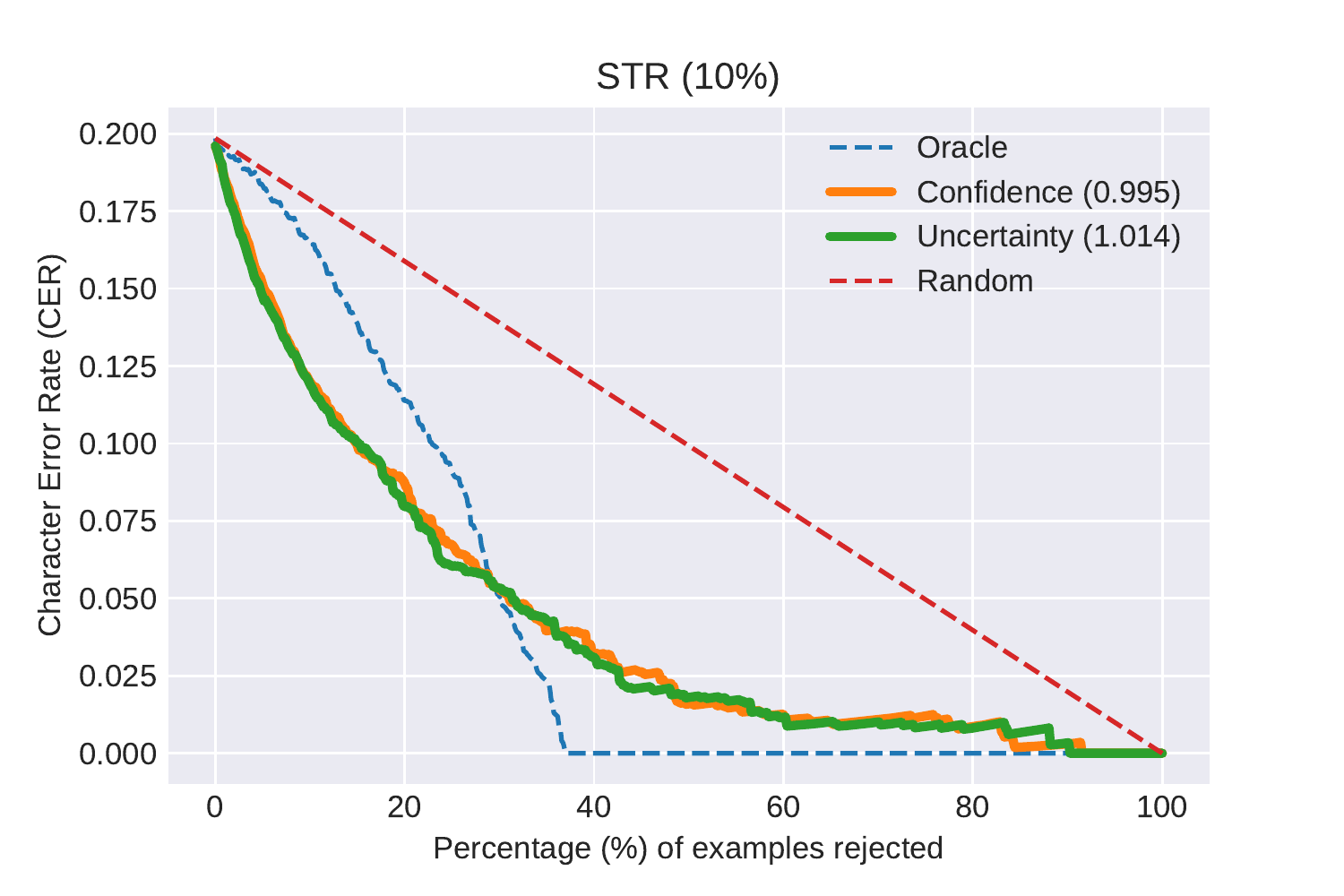}\hfill
    \includegraphics[width=0.24\textwidth]{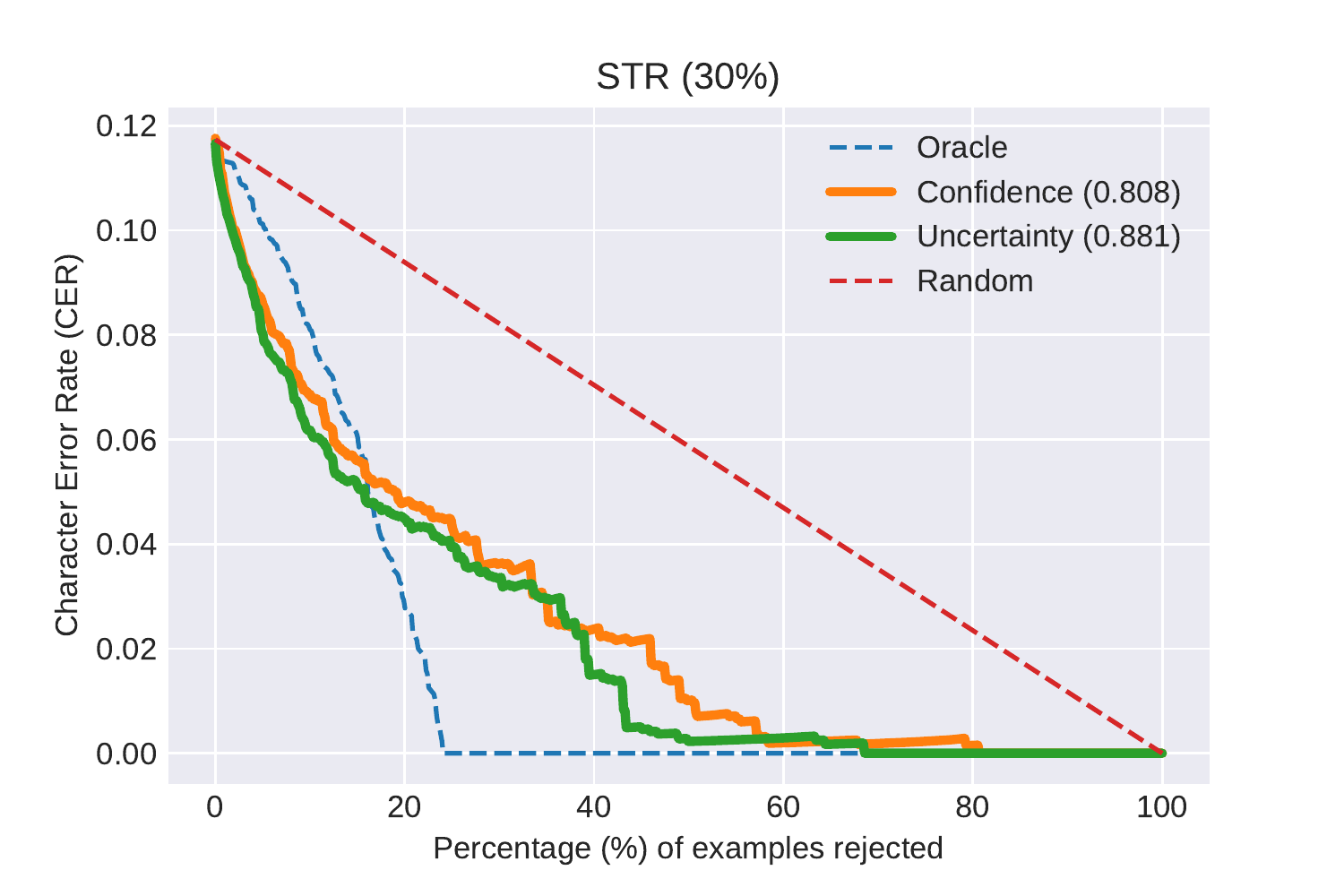}\hfill
    \includegraphics[width=0.24\textwidth]{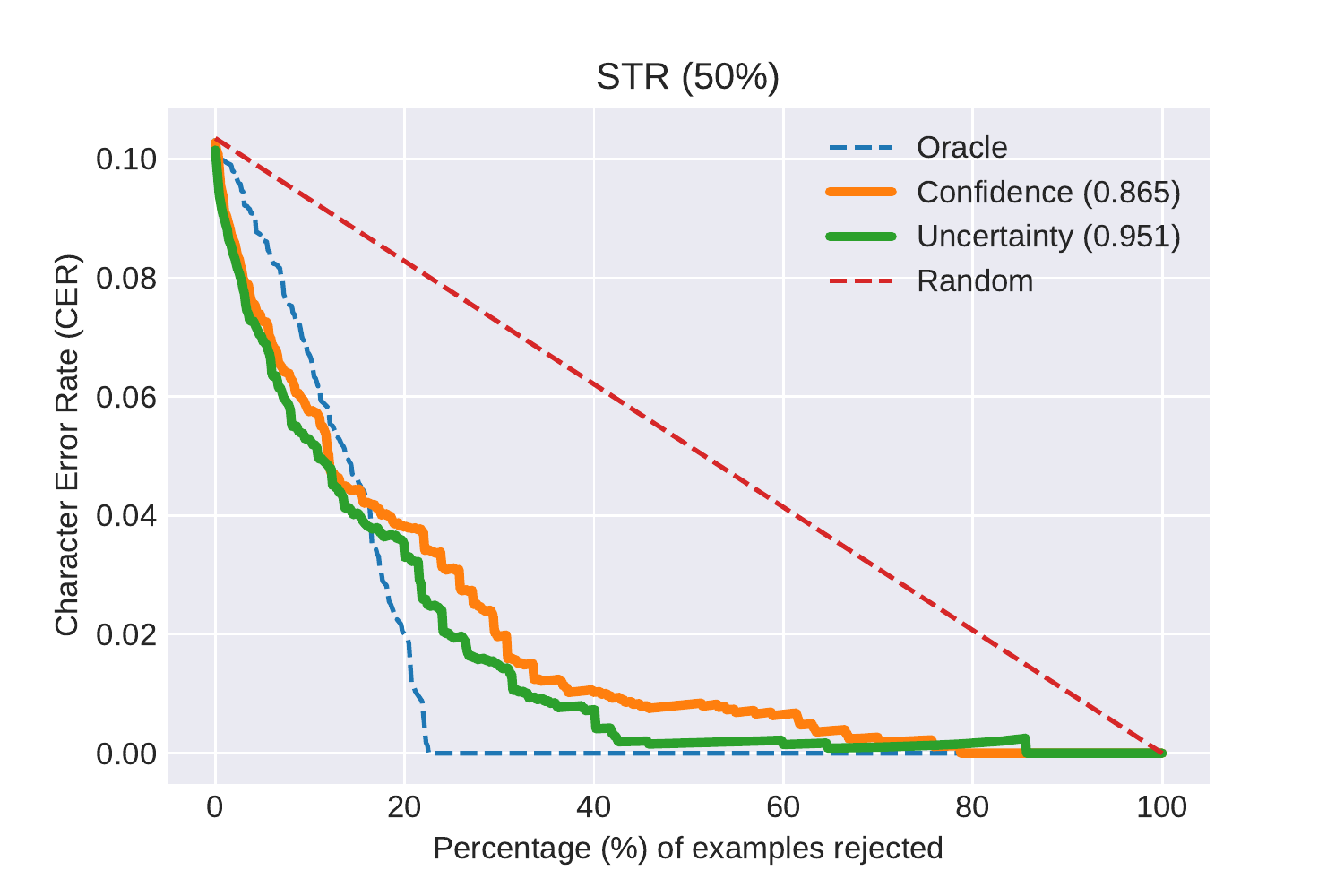}
    \caption{Prediction Rejection Curves w.r.t Character Error Rate (CER). Values in parenthesis in the legend field represent the Prediction Rejection Ratio (PRR) of the corresponding curve.}\label{fig:rej_curves_2}
\end{figure*}

\begin{figure*}[ht]
\centering
    \includegraphics[width=.16\textwidth]{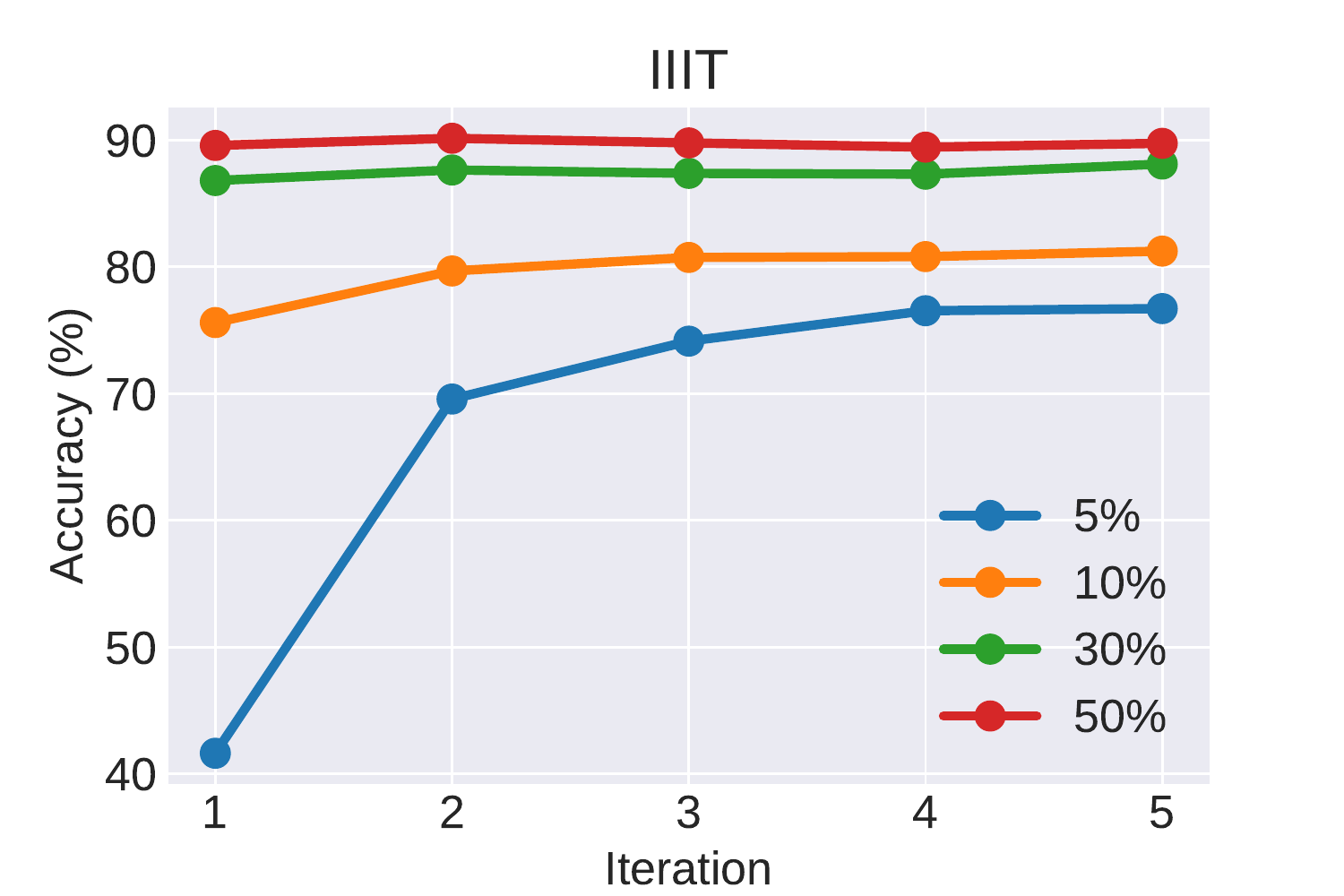}\hfill
    \includegraphics[width=.16\textwidth]{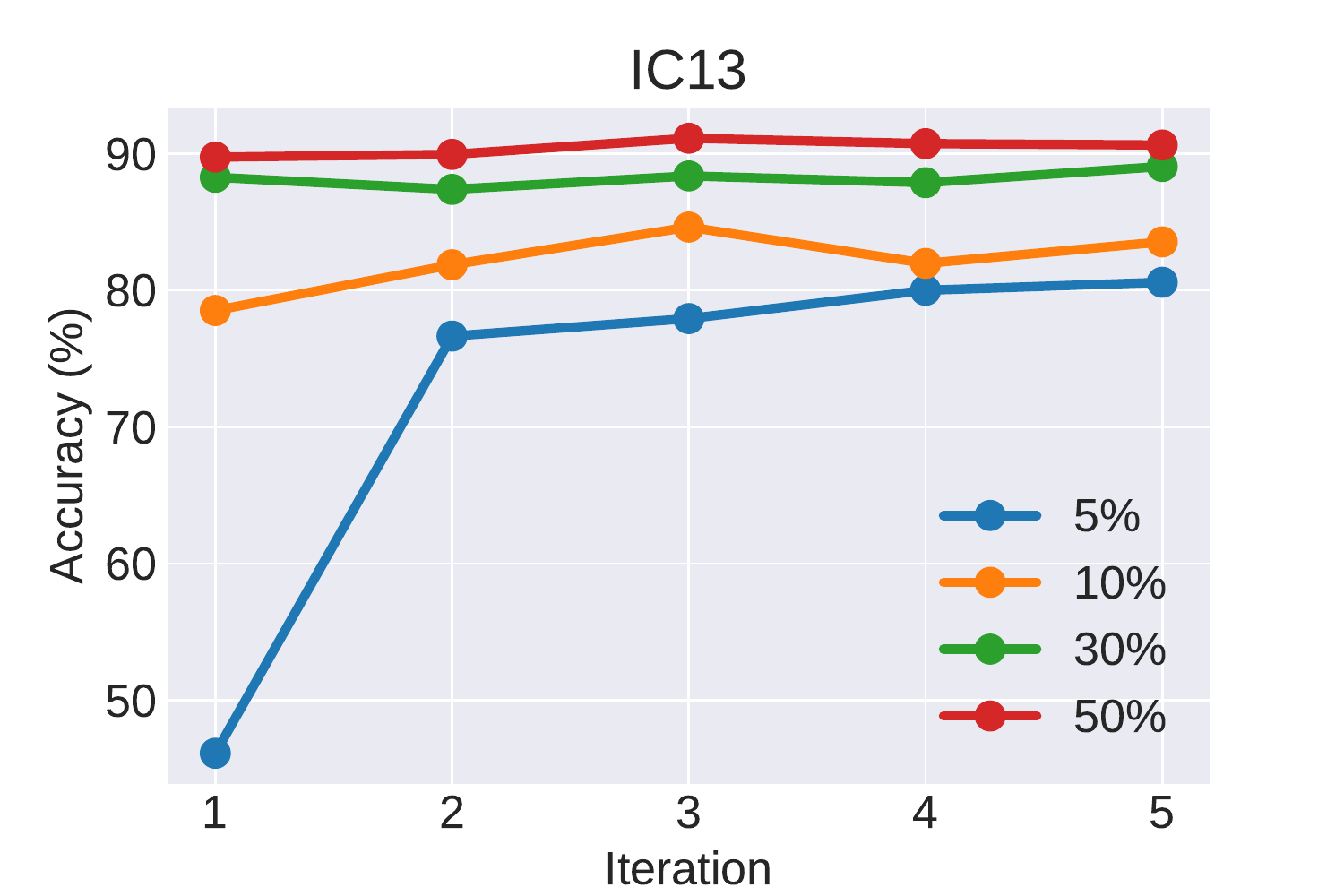}\hfill
    \includegraphics[width=.16\textwidth]{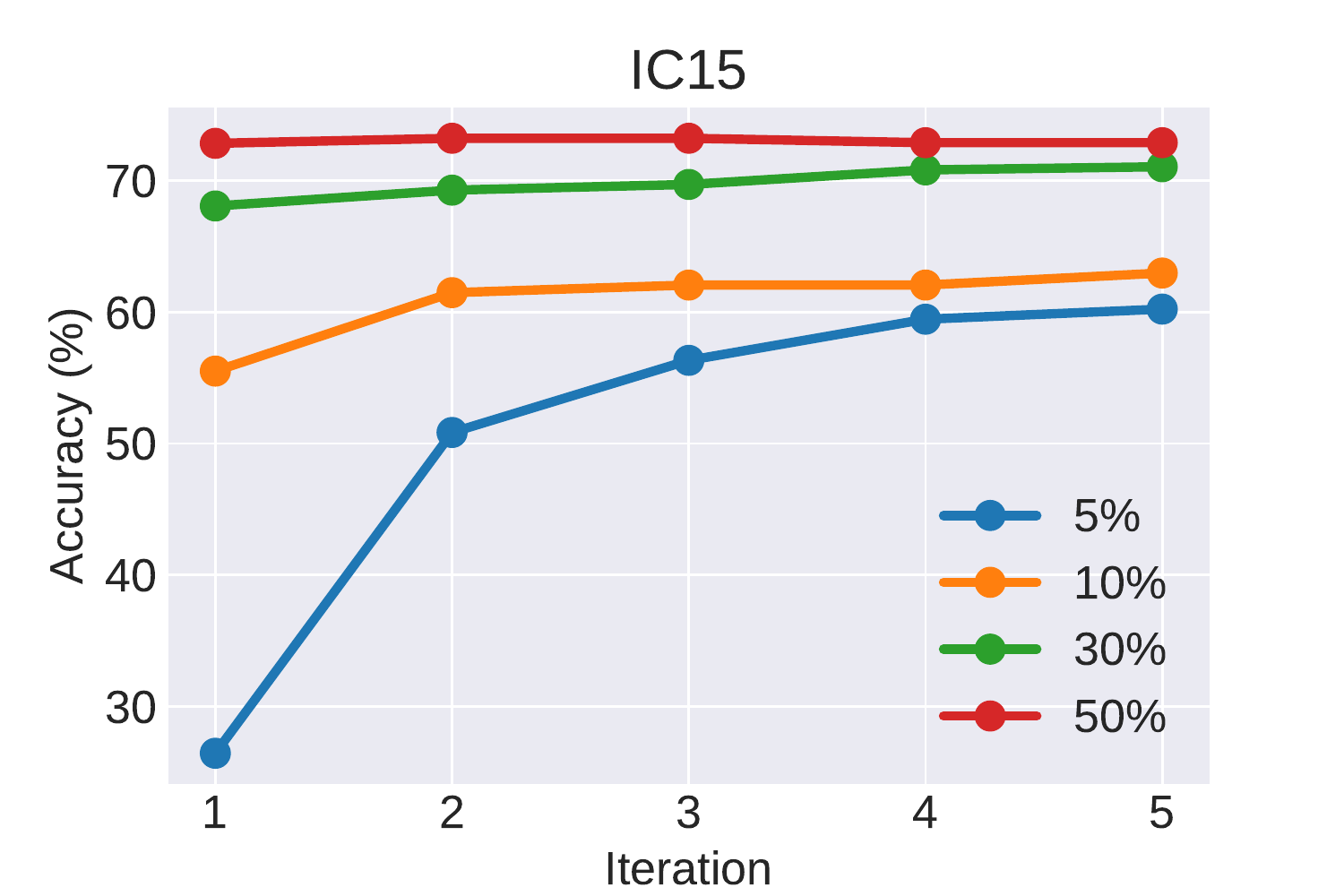}\hfill
    \includegraphics[width=.16\textwidth]{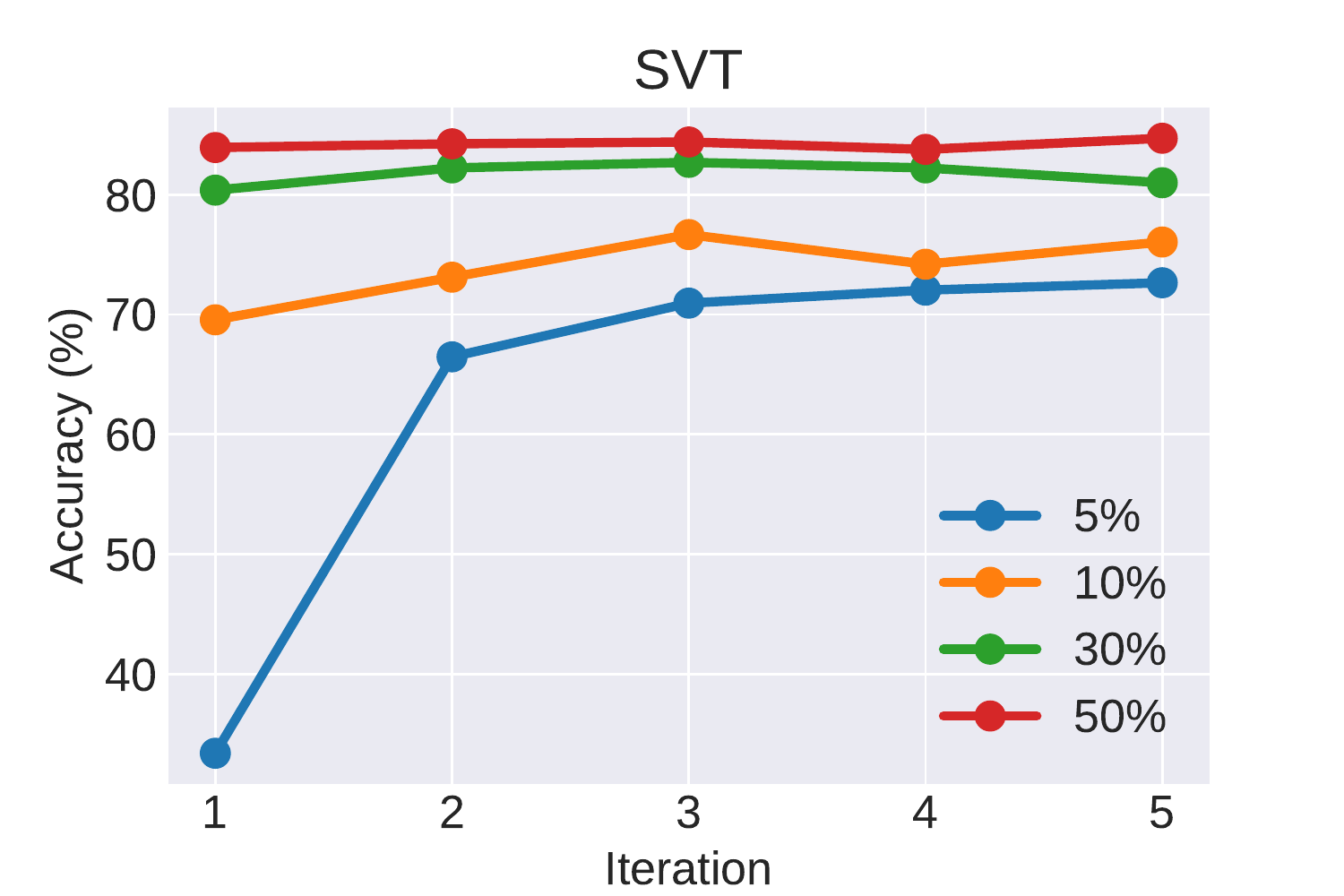}\hfill
    \includegraphics[width=.16\textwidth]{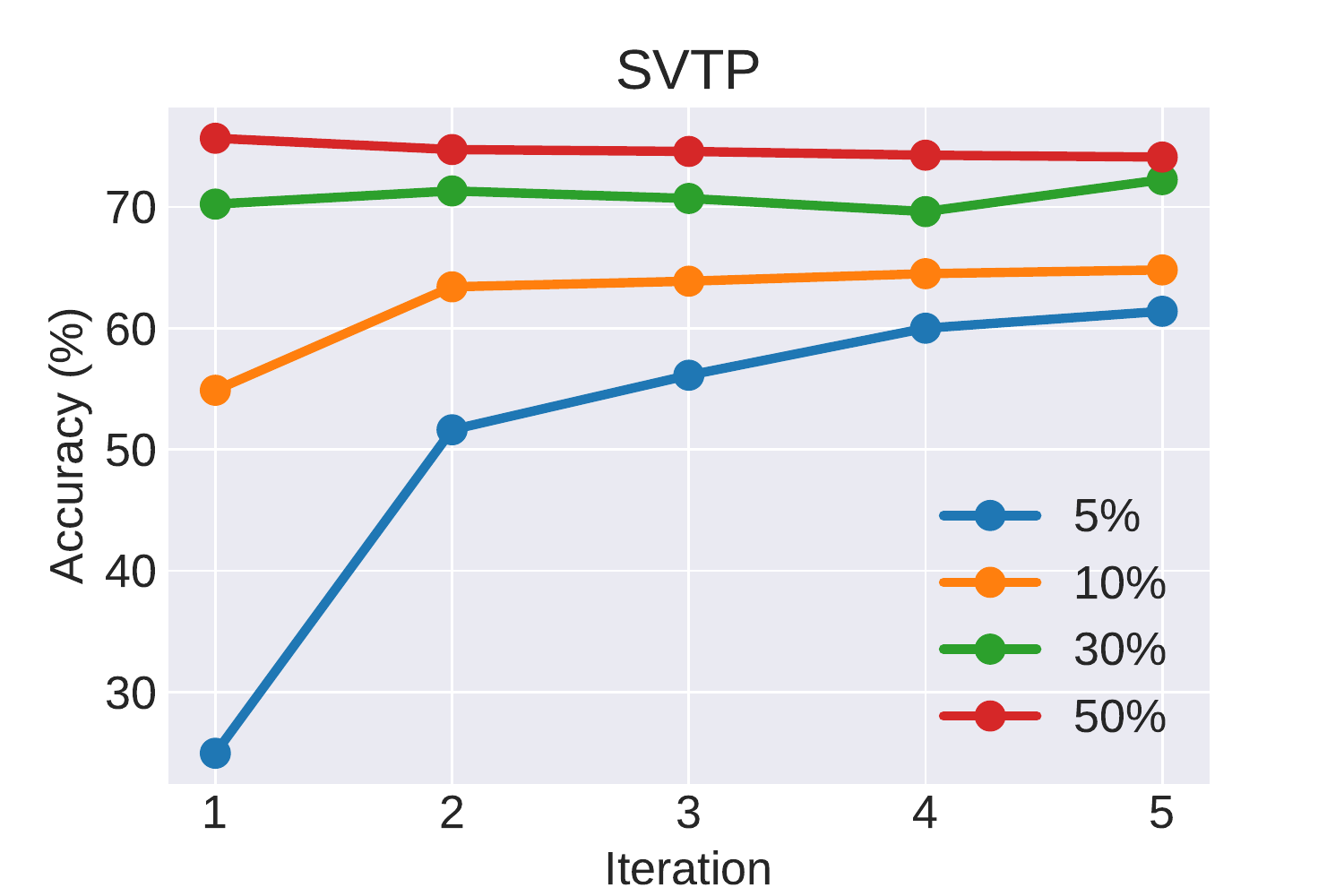}\hfill
    \includegraphics[width=.16\textwidth]{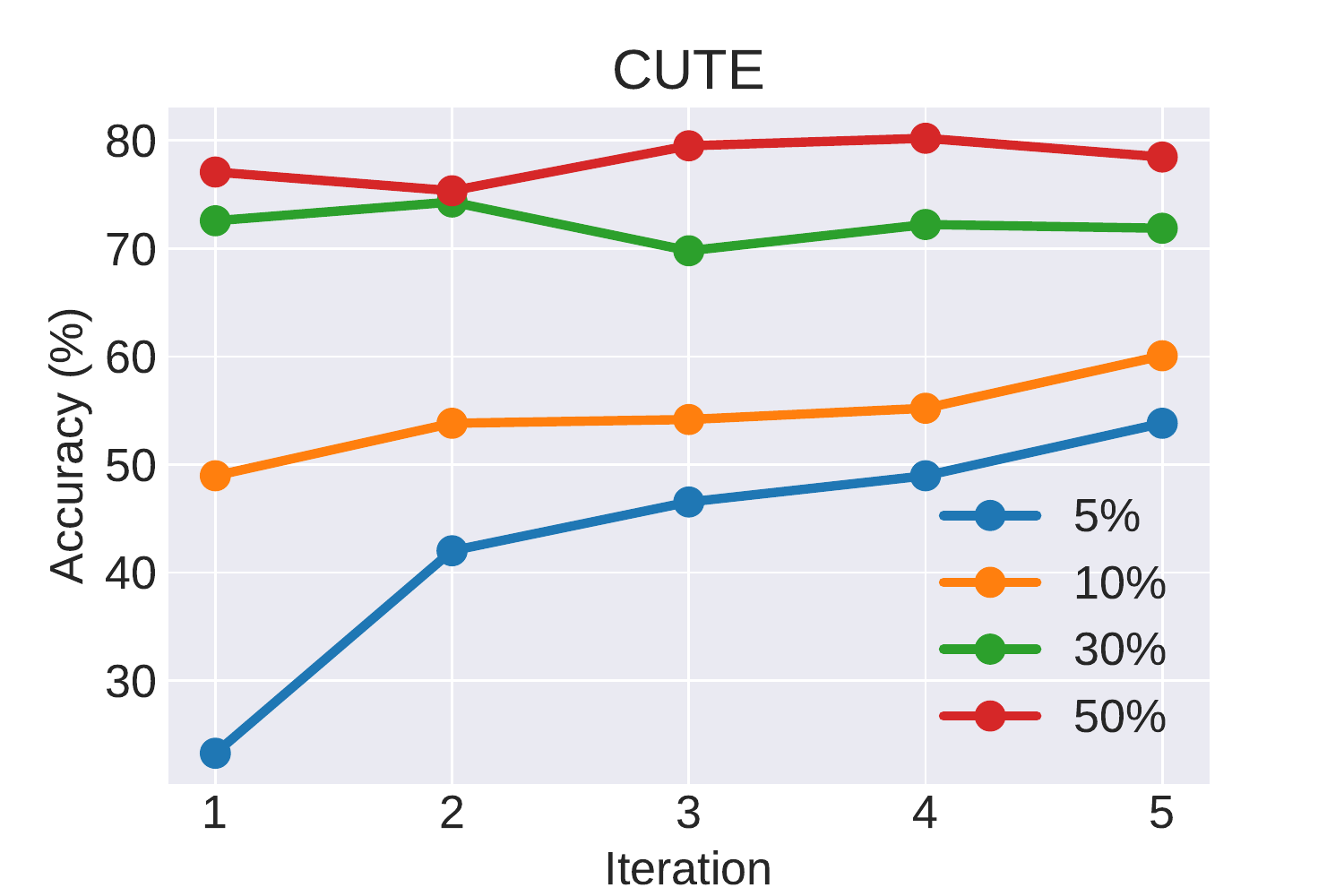}

    \includegraphics[width=.16\textwidth]{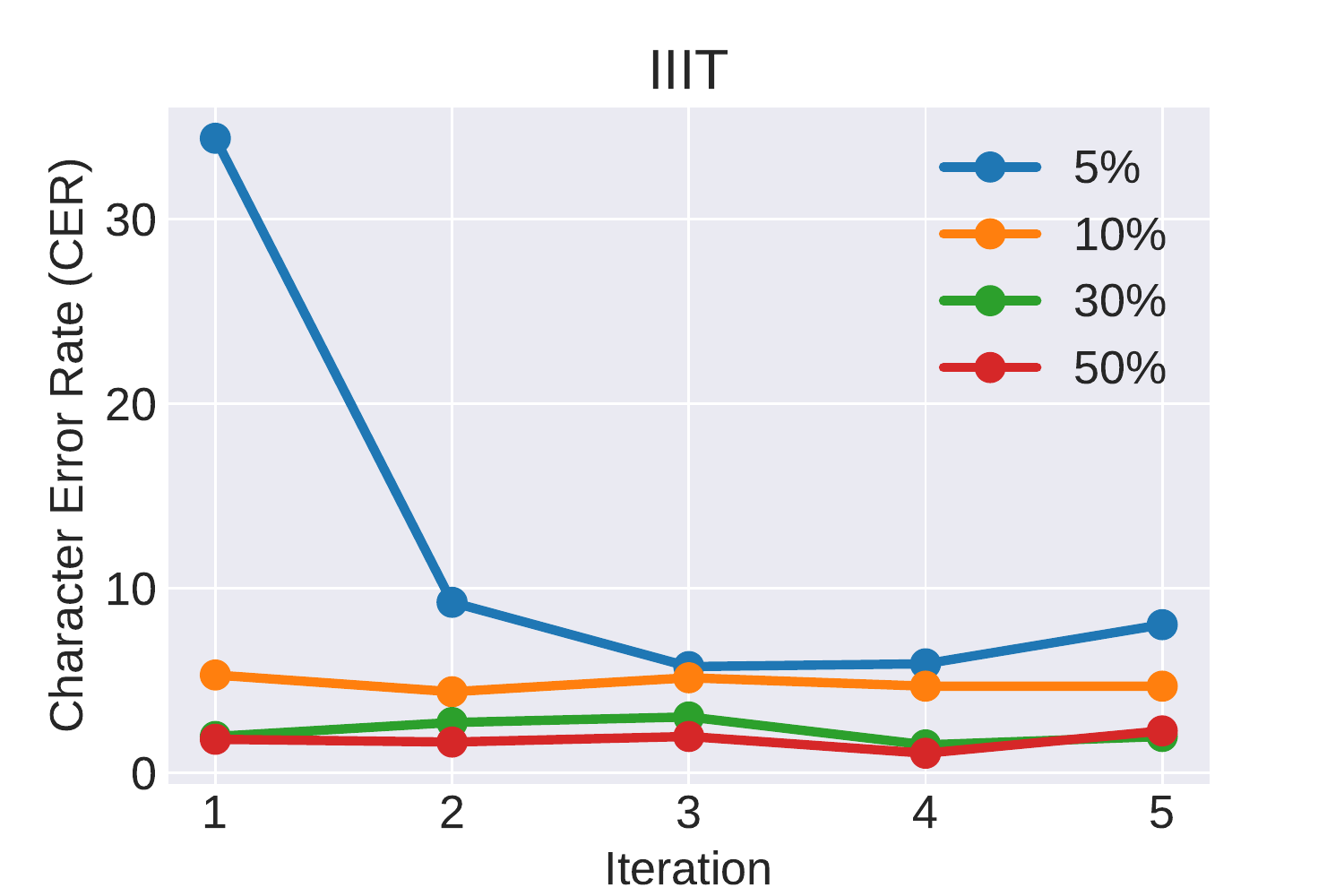}\hfill
    \includegraphics[width=.16\textwidth]{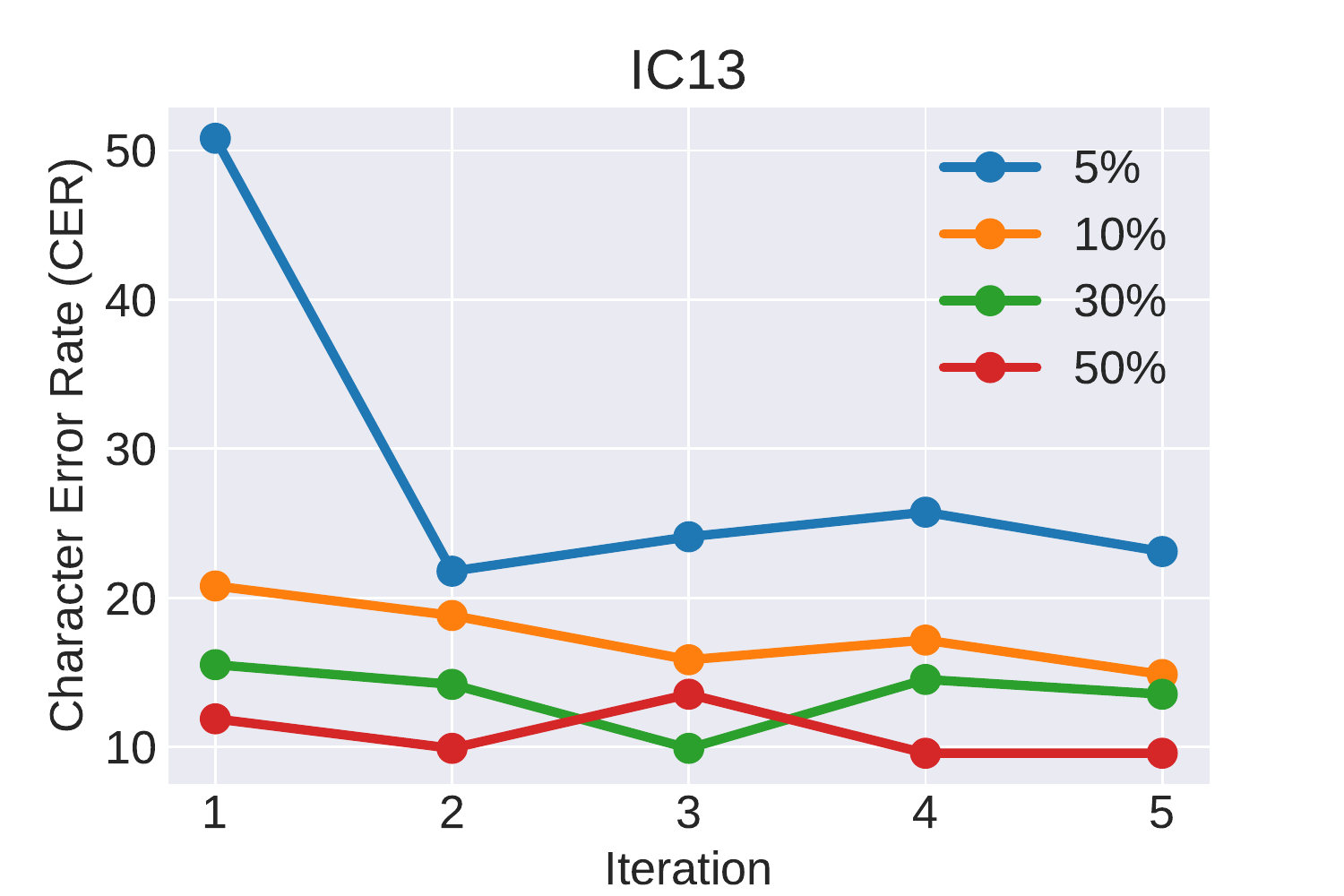}\hfill
    \includegraphics[width=.16\textwidth]{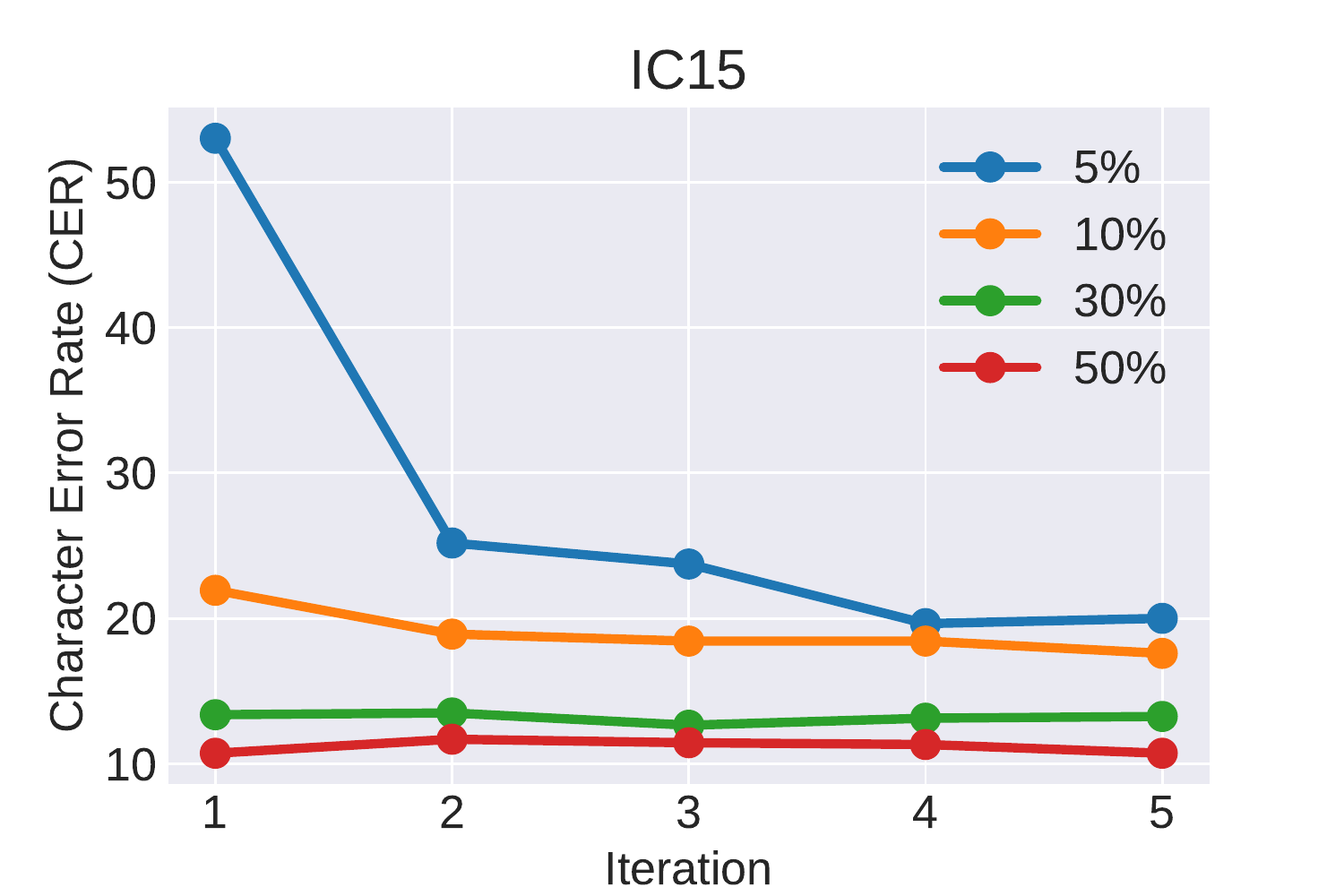}\hfill
    \includegraphics[width=.16\textwidth]{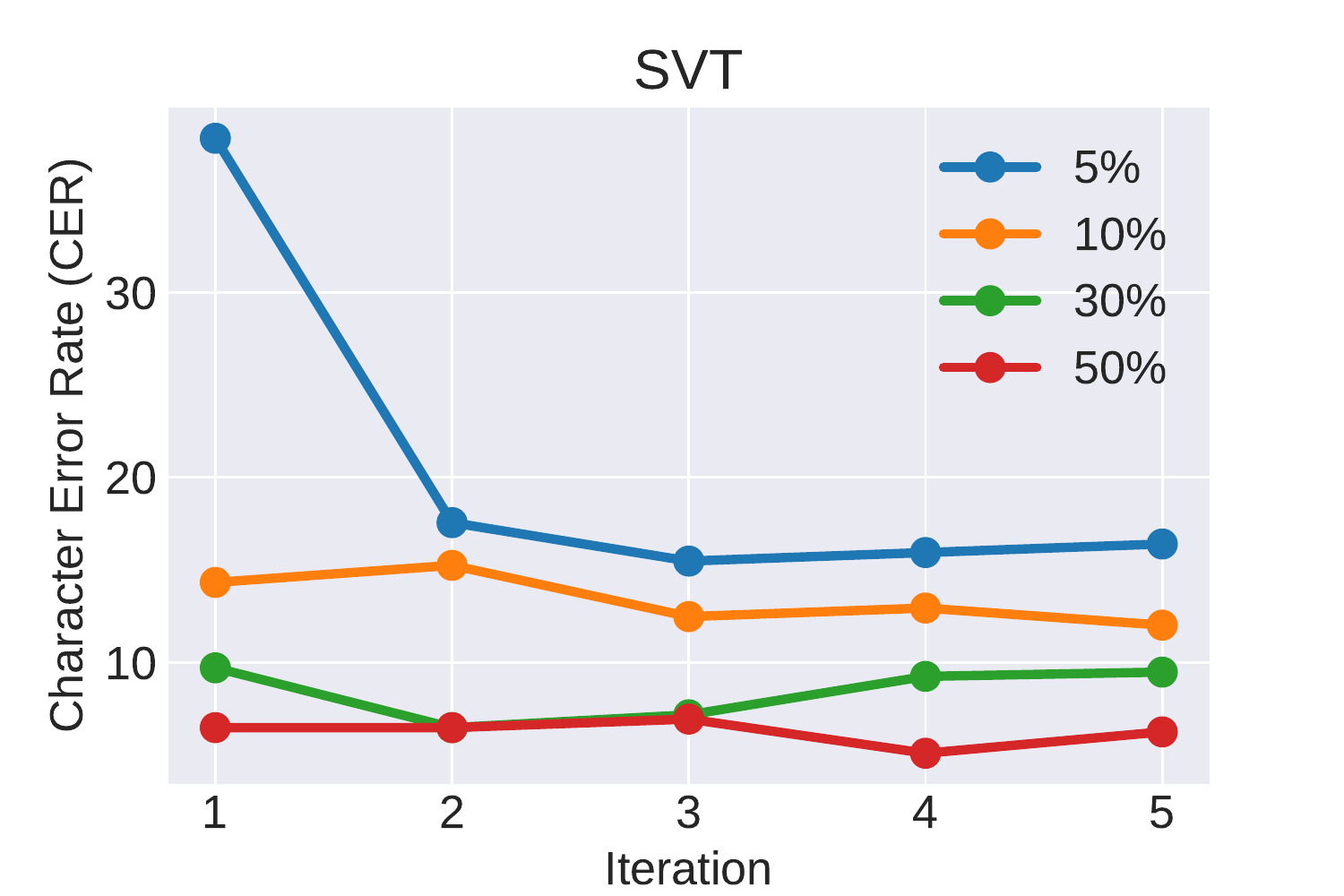}\hfill
    \includegraphics[width=.16\textwidth]{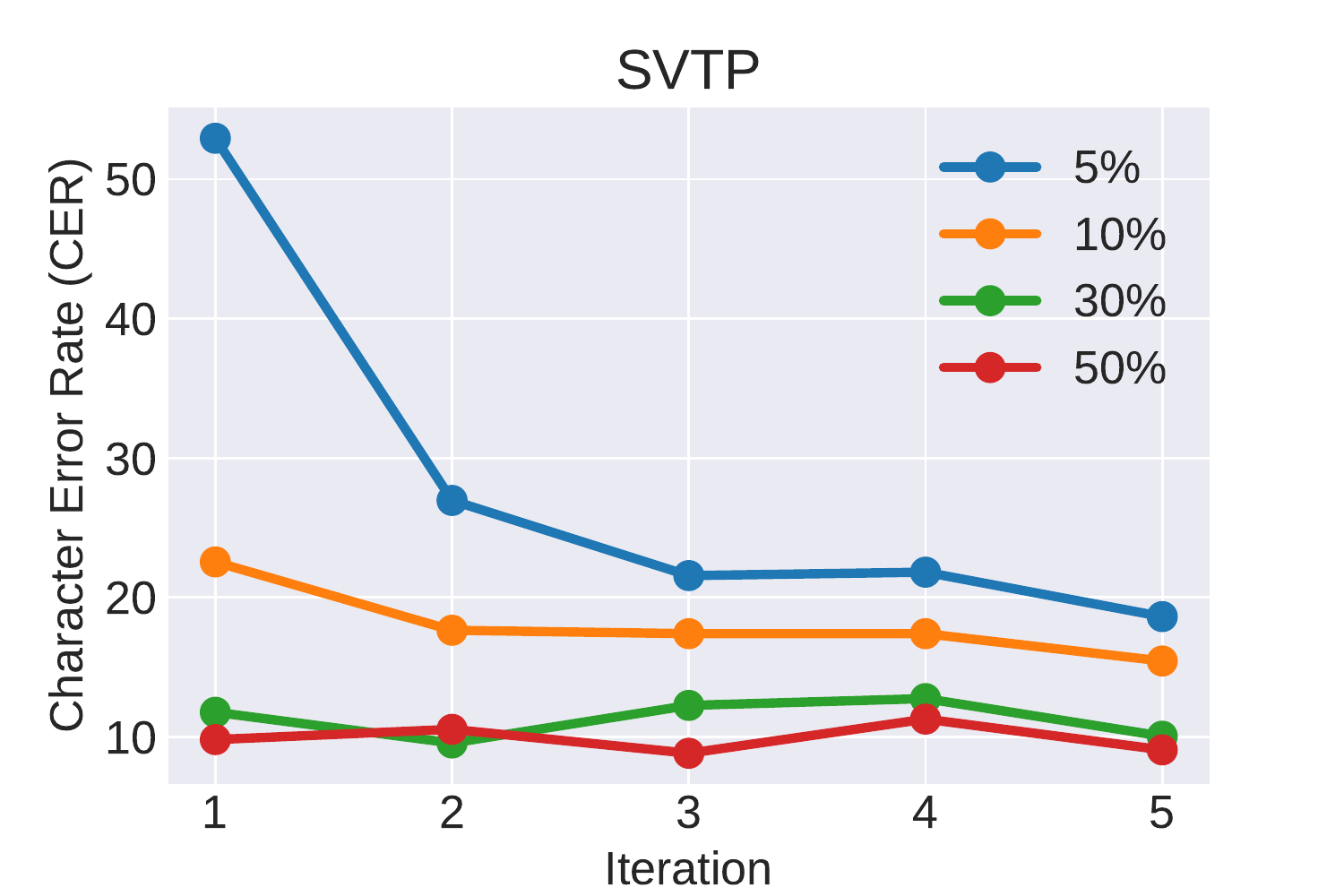}\hfill
    \includegraphics[width=.16\textwidth]{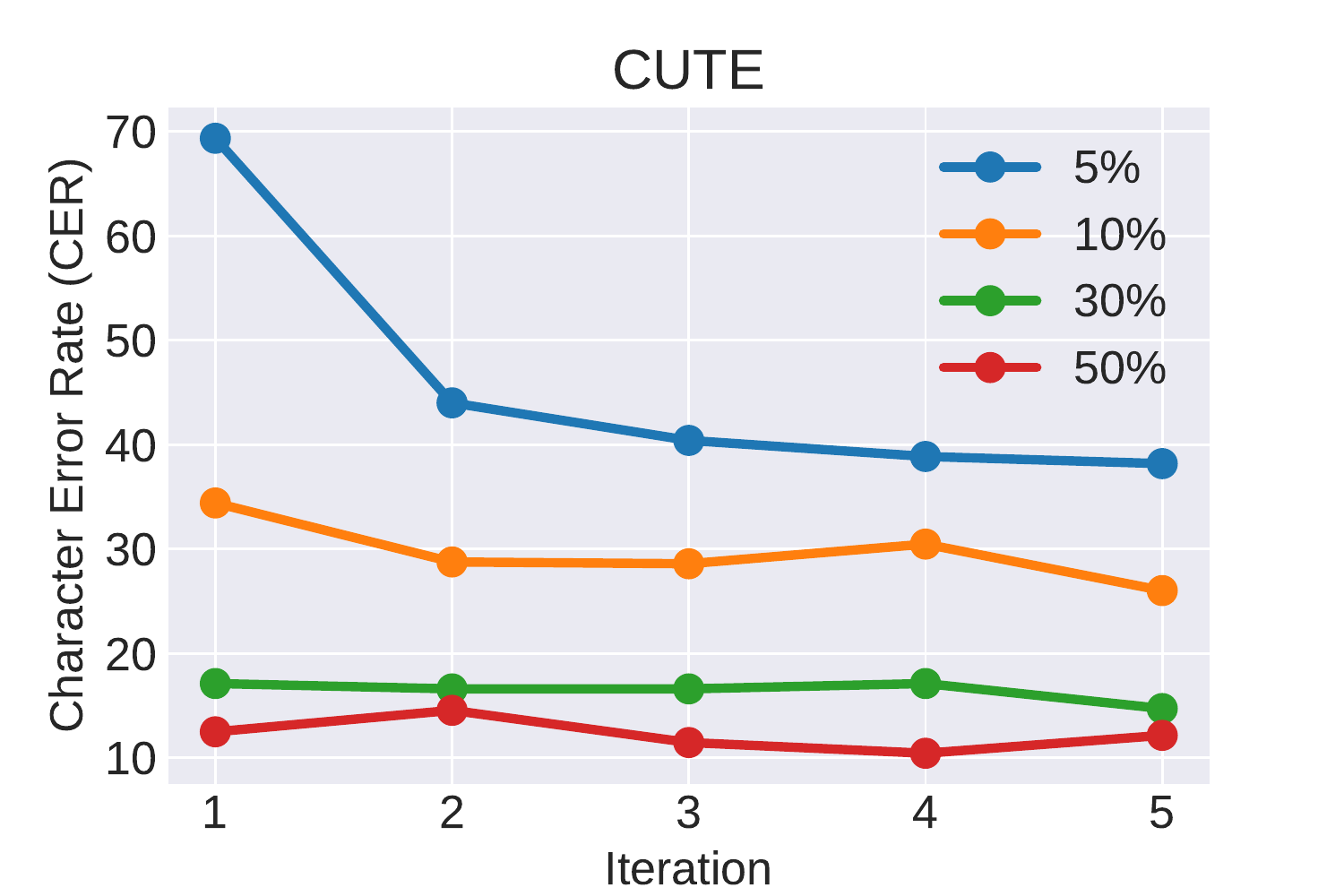}
    \caption{Iteration-wise metric trends of the pseudo-labeling based semi-supervised learning methodology with randomly initialized weights of individual benchmark scene-text datasets. }\label{fig:metric_trends_2}
\end{figure*}

\setlength{\tabcolsep}{4pt}
\begin{table}[ht]
\centering
\caption{Configuration of the system used to train the models}\label{table:configuration}
\resizebox{0.99\textwidth}{!}{\begin{tabular}{|c|cl|c|l|}
\hline\noalign{\smallskip}
\textbf{RAM} & \multicolumn{2}{c|}{\textbf{CPU}}                              & \textbf{VRAM}  & \multicolumn{1}{c|}{\textbf{GPU}}             \\ \noalign{\smallskip}\hline\noalign{\smallskip}
251 GB       & \multicolumn{2}{c|}{Intel Core i9-10940X} & 11$\times$4 GB & Nvidia RTX 2080Ti \\ \noalign{\smallskip}\hline
\end{tabular}}
\end{table}

\begin{table*}[ht]
\centering
\caption{Word level accuracy (Acc \%) using all the labeled data and additional unlabelled data.}
\label{table:results}
\resizebox{0.95\textwidth}{!}{\begin{tabular}{|l|ccccccr|}
\hline
\textbf{Method}                              & \textbf{IIIT5K} & \textbf{SVT} & \textbf{IC13} & \textbf{IC15} & \textbf{SVTP} & \textbf{CUTE80} & \multicolumn{1}{c|}{\textbf{Total}} \\ \hline
Supervised Baseline                & 92.27           & 85.63        & 92.22         & 74.96         & 75.97         & 83.68           & 85.36                               \\
SeqCLR (All-to-instance)            & 92.23           & 86.86        & 91.43         & 76.97         & 77.99         & 82.99           & 86.02                               \\
SeqCLR (Frame-to-instance)          & 91.13           & 87.79        & 92.02         & 77.85         & 78.30          & \textbf{86.11}           & 86.13                               \\
SeqCLR (Window-to-instance)         & 91.23           & 87.64        & \textbf{93.01}         & \textbf{77.90}          & 80.16         & 85.76           & 86.44                               \\ \hline
\textbf{Ours}                                & 92.73           & \textbf{88.56}        & 92.22         & 76.87         & 77.98         & 84.37           & 86.50                                \\
\textbf{Ours w/ SeqCLR (All-to-instance)}    & 92.33           & 87.17        & 91.82         & 77.85         & 78.92         & 85.42           & 86.55                               \\
\textbf{Ours w/ SeqCLR (Frame-to-instance)}  & 92.83           & 86.71        & 92.61         & 77.36         & 79.23         & \textbf{86.11}           & 86.73                             \\
\textbf{Ours w/ SeqCLR (Window-to-instance)}  & \textbf{93.13}           & 86.71        & 91.72         & 76.83         & \textbf{81.40}          & 85.90            & \textbf{86.76}   \\
\hline
\end{tabular}}

\end{table*}

\section{Training and Evaluation Details}
To train the models we use the AdaDelta \cite{adadelta} optimizer with a learning rate of $1$ and a decay rate of $\rho = 0.95$. Furthermore, in total we perform 4 pseudo-label based fully-supervised re-training of the model in a bootstrapped fashion after the initial fully-supervised training with the partially labeled dataset. For each of the fully-supervised training, we train the model for 100K iterations with a batch size of 192. Furthermore, for stable training we use gradient clipping of magnitude 5. Moreover, we use He's method to initialize all parameters. All the models were trained on a single GPU on a server with the configuration described in \ref{table:configuration}. Algorithm \ref{alg:one} describes the the pseudo-label assignment and selection of the unlabelled data samples, that return $\mathcal{D}_{train}$ updated with the pseudo-labeled samples.

\begin{algorithm*}[ht]
\caption{Pseudo-label assignment and selection of unlabelled data samples at the end of $I$-th training iteration for the subsequent iteration of supervised training.}\label{alg:one}
\KwData{$\mathcal{D}_{train}$, $\mathcal{D}_{u}$, $\boldsymbol{\theta}_{I}$, $\tau$}
\KwResult{$\mathcal{D}_{train}$}
$N_{u}$ = Number of samples in $\mathcal{D}_{u}$.\\
$\mathcal{B}_{i}$ = Hypotheses set for $i$-th sample.\\
$\tilde{Y}_{i}^{u}$ = Pseudo label for $i$-th sample.\\
$i = 1$ \;
\While{$i \leq N_{u}$}{
    $\mathcal{B}_{i} \gets \bsi{(f_{\boldsymbol{\theta}_{I}}(X_{i}^{u}))}, X_{i}^{u} \in {\mathcal{D}_{u}} $\Comment*[r]{Deterministic Inference}
    $\tilde{Y}_{i}^{u} \gets \operatorname*{arg\,max}_{Y^{(b)}_{i}}\{P(Y^{(b)}_{i}|X_{i}^{u};\boldsymbol{\theta}_{I})\}_{b=1}^{B}, Y_{i}^{(b)} \in \mathcal{B}_{i}$\Comment*[r]{Pseudo-Label assignment}
    
    Compute $\mathcal{U}({X_{i}^{u}, \mathcal{B}_{i})}$ using (6) from the main script\Comment*[r]{Stochastic Inference}
    
    \If{$\mathcal{U}({X_{i}^{u}, \mathcal{B}_{i})} \leq \tau$}{
        $\mathcal{D}_{train} \gets \mathcal{D}_{train} \cup \{X_{i}^{u}, \tilde{Y}_{i}^{u}\}$ \;
    }
    $i$+=$1$\;
}
\end{algorithm*}

Also, MC-Dropout \cite{mcdropout} is notorious for being computationally inefficient since it requires passing every input to each of the sampled model to compute the uncertainty. However, in our implementation we utilize an efficient batch implementation that can easily replace the vanilla Dropout layers in PyTorch\footnote{\url{https://blackhc.github.io/batchbald_redux/consistent_mc_dropout.html}}. The efficient dropout layer keeps a set of dropout masks fixed while scoring the pool set and exploit batch parallelization for scalibility \cite{kirsch2019batchbald}, thus, alleviating the need to pass the input multiple times and the explicit sampling of the models in the ensemble, thus making the system both memory and computationally efficient.

To train the handwriting recognition model we utilize the training splits of the IAM \cite{IAM} and the CVL \cite{CVL} and for the scene-text recognition model, contrary to the previous works \cite{seqCLR,deeptext} that use synthetic datasets \cite{ST,MJ}, we use a combination of multiple real scene-text datasets, following the work in \cite{RealSTR_data}, for training, that include: IC15 \cite{ICDAR15}, IC13 \cite{ICDAR13}, IIIT \cite{IIIT}, SVT \cite{SVT}, SVTP \cite{SVTP}, CUTE \cite{CUTE}, COCO-Text \cite{veit2016coco}, RCTW \cite{RCTW}, Uber-Text \cite{UberText}, ArT \cite{Art}, MLT \cite{MLT19}, and ReCTS \cite{ReCTS} consolidating a total of 276k processed images in the training set\footnote{Preprocessed scene-text dataset with the training, validation, and test splits are made available by the authors of \cite{RealSTR_data} at: \url{https://github.com/ku21fan/STR-Fewer-Labels}}.


The models are evaluated on the IAM \cite{IAM} and CVL \cite{CVL} test sets for handwriting recognition. For scene-text recognition we benchmark on six scene-text datasets: IC13 \cite{ICDAR13}, IC15 \cite{ICDAR15}, IIIT \cite{IIIT}, SVT \cite{SVT}, SVTP \cite{SVTP}, CUTE \cite{CUTE}. For comparison, we also determine the total accuracy, which is the accuracy of the six benchmark datasets combined. Specifically, for scene-text evaluation, the accuracy is calculated only on alphabet and digits, after removing non-alphanumeric characters and normalizing alphabet to lower case. Furthermore, we execute three trials with different seed values for the experiments and report the averaged accuracies.

\section{Additional Results}

\paragraph{}
In Figure \ref{fig:rej_curves_2}, we visualize the prediction rejection curves w.r.t to the character error rate (CER) of the baseline text recognition model trained on different portions of labeled data on the handwriting and the scene-text datasets. 

\paragraph{}
In Figure \ref{fig:metric_trends_2}, we show our vanilla  PL-SSL method's performance on word prediction accuracy and CER at the end of each supervised training iteration, starting with different portions of labeled training dataset, for each individual scene-text benchmarks.

\paragraph{}
Moreover, We conduct experiments with all the text images in the labeled set (276K instances) and the text instances from the TextVQA dataset \cite{singh2019towards} (463K instances) as the unlabeled set, in Table \ref{table:results}. We found our methods to give on par and in some cases better performance to SeqCLR \cite{seqCLR}.
